\documentclass[11pt]{article}

\usepackage[preprint]{acl}
\usepackage{xurl}
\usepackage{times}
\usepackage{latexsym}
\usepackage[T1]{fontenc}
\usepackage[utf8]{inputenc}
\usepackage{microtype}
\usepackage{inconsolata}
\usepackage{graphicx}
\usepackage{amsmath,amssymb}
\usepackage{algorithm}
\usepackage{algorithmic}
\usepackage{booktabs}
\usepackage{multirow}
\usepackage{fvextra}
\usepackage{xcolor} 
\usepackage[misc]{ifsym}
\newcommand{\ours}[1]{\textsc{RIVER}}

\title{Beyond Final Outcomes: Quantifying Claim-Level\\Information Dynamics in Multi-Agent LLM Interaction}

\author{
  Haocheng Jiang \\
  Xi'an Jiaotong University \\
  \texttt{jhc060321@stu.xjtu.edu.cn}
  \And
  Herun Wan \\
  Xi'an Jiaotong University \\
  \texttt{wanherun@stu.xjtu.edu.cn}
  \AND
  Yongyuan Peng \\
  Xi'an Jiaotong University \\
  \texttt{yongyuan16625@gmail.com}
  \And
  Minnan Luo\textsuperscript{\Letter}\thanks{Corresponding author.} \\
  Xi'an Jiaotong University \\
  \texttt{minnluo@xjtu.edu.cn}
}

\begin{document}
\maketitle

\begin{abstract}

When LLM-powered agents interact, information is repeatedly shared, expanded, and forgotten, yet multi-agent evaluation typically focuses on final task outcomes. We study such interaction as claim-level information dynamics and propose \ours{}, which decomposes agent outputs into atomic claims and evaluates three complementary dimensions: \textbf{convergence}, \textbf{exploration}, and \textbf{retention}. \textbf{Convergence} captures cross-agent claim correspondence and stance alignment; \textbf{exploration} captures semantic breadth and extension beyond the source; and \textbf{retention} tracks source preservation and the persistence of new claims across stages and lineages. We evaluate five interaction strategies on 40 controlled news-generation cases. Repeatedly pooled interaction promotes convergence but narrows the information space, whereas distributed relay yields broader, less shared information. Further comparisons show that method-specific constraints increase exploration while weakening convergence and retention. Our findings suggest that collective information dynamics are better explained by underlying information-flow mechanisms than by simple downstream performance.\footnote{Code and data are available on \href{https://github.com/fzj-2023/RIVER}{[GitHub Repository]}.}
\end{abstract}

\section{Introduction}\label{sec:introduction}

\begin{figure*}[t]
  \centering
  \includegraphics[width=\textwidth]{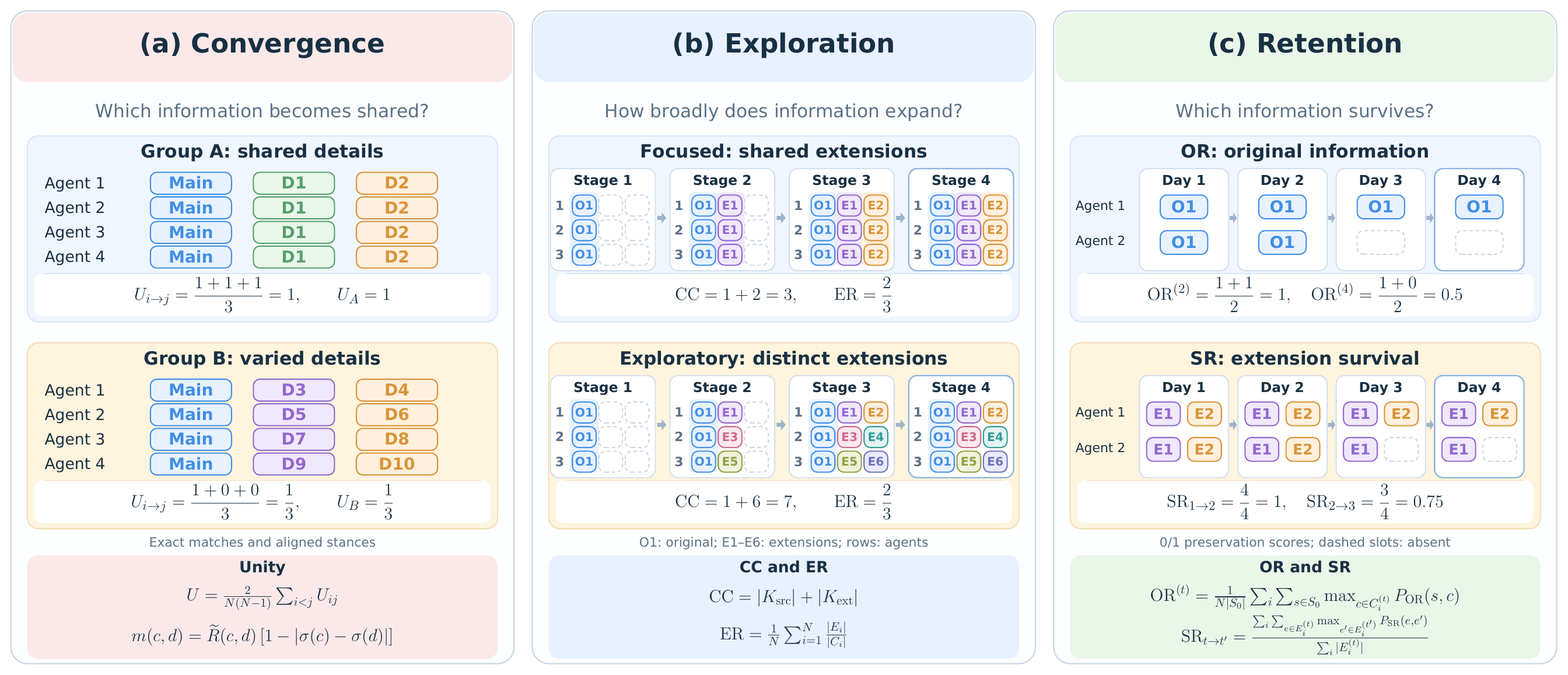}
  \caption{\textbf{Overview of claim-level information-flow evaluation along three complementary dimensions.} \textbf{(a)} \textbf{Convergence} measures whether agents express the same specific information with aligned stances, illustrated by Informational Unity (U). \textbf{(b)} \textbf{Exploration} characterizes how broadly the group expands its information space, using Unique Claim Count (CC) and Extension Rate (ER). \textbf{(c)} \textbf{Retention} tracks which information persists over time, illustrated by Original News Retention (OR) and Survival Rate (SR).}

  \label{fig:information_overview}
\end{figure*}

Large language models (LLMs) increasingly interact as groups. Multiple LLM-powered agents can propose ideas, exchange information, revise beliefs, and synthesize conclusions over several rounds \citep{du2024debate}. The interactions are organized by a strategy that determines what each agent observes \citep{hong2024metagpt}, whose information it receives \citep{shen2025propagation}, and how previous content is summarized \citep{park2023generativeagents}. In summary, they determine how information flows among the group.

Despite growing interest in multi-agent LLM systems, relatively little attention is paid to what happens to information during these interactions. Two groups may reach the same final judgment while expressing substantially different shared information states. 
Beyond shared information, agents may introduce the same new claims or distinct ones, thereby expanding the information space to different extents. In addition, both source information and new claims may persist or disappear during interaction.

Figure~\ref{fig:information_overview} presents the characteristics, where a claim may be repeatedly exposed until it becomes shared across agents, expand into multiple new claims, or be introduced once and disappear. Final accuracy \citep{du2024debate}, task successes \citep{liu2024agentbench,zhu2025multiagentbench}, or output similarity \citep{zhang2020bertscore} cannot distinguish these processes. 
These patterns show that agreement at the decision level does not imply agreement at the information level, and that final outcomes alone do not adequately characterize information changes during interaction. The multi-agent interaction should be considered as a process of collective information dynamics, in which information is continuously introduced, exposed, reproduced, transformed, retained, and lost.
Understanding multi-agent interaction therefore requires evaluating which information becomes shared, what new information is introduced, and which claims survive or disappear.

To this end, we propose \ours{}\footnote{\ours{} stands for \textbf{\underline{R}}eveal \textbf{\underline{I}}nformational con\textbf{\underline{V}}ergence, \textbf{\underline{E}}xploration and \textbf{\underline{R}}etention.}, a claim-level framework for measuring these properties of multi-agent collaboration strategy. It evaluates the information flow from three complementary dimensions: \textbf{convergence}, \textbf{exploration}, and \textbf{retention}. \textbf{Convergence} asks whether agents come to express the same specific information and hold aligned views towards it, which we measure with Informational Unity (U). \textbf{Exploration} characterizes how broadly the group expands its information space, measured through the number of semantically distinct claims (Unique Claim Count, CC) and the proportion of information introduced beyond the initial source (Extension Rate, ER). \textbf{Retention} captures which information persists as interaction progress, including preservation of the original source (Original News Retention, OR), survival of newly introduced claims across stages (Survival Rate, SR), and the duration of their traceable lineages (Claim Survival Length, SL). In summary, \ours{} provides a unified view of what information becomes shared, what new information is explored, and what information survives. We additionally validate Informational Unity against human-derived signals of information sharing and agreement.

We employ \ours{} to explore five representative interaction strategies in controlled multi-agent news discussion, including Free, NGT, Delphi, Policy Delphi, and Brainwriting. Across 40 news instances, all groups operate under the same generation settings while the interaction strategies determine how information is exposed, revised, and propagated. For the strategies except Free, we further separate the core interaction framework from the additional method-specific constraints by comparing a Framework setting with the corresponding Full setting. It allows us to examine whether different strategies produce different information dynamics, and whether these differences arise from the underlying communication structure or from the behavioral constraints layered on top of it.

Our results reveal systematic differences in how interaction strategies organize information. Repeatedly pooled interaction tends to increase convergence while concentrating the information space, whereas Brainwriting's distributed relay maintains broader but less shared information. Meanwhile, by comparing Framework and Full settings, we find that interaction strategies do not behave as single, indivisible mechanisms. Across all four strategies, adding specific constraints increases information exploration but weakens convergence and retention. Groups express more distinct and source-extending claims, while sharing less information and preserving less source and extension content over time. These patterns suggest that collective information dynamics are better understood through the underlying information-flow structure and behavioral constraints than through downstream performance.

\section{Related Work}

\paragraph{LLM-based Multi-Agent Systems.}
Recent advances in large language models (LLMs) have enabled LLM-based agents to perform complex tasks through reasoning \citep{wei2022chainofthought}, planning \citep{park2023generativeagents}, and interaction with the environment \citep{yao2023react}. For example, these agents have been applied to question answering \citep{yao2023react}, machine translation \citep{wu2024transagents}. Multi-agent systems extend these capabilities through communication, role assignment, and mutual feedback among agents \citep{li2023camel,hong2024metagpt}. These systems support reasoning and decision-making \citep{du2024debate}, software development and content generation \citep{qian2024chatdev}, and misinformation detection \citep{wan2024dell,xiong2025delphiagent}. They are also used to simulate news propagation \citep{liu2025fuse} and social interactions within groups \citep{park2023generativeagents}. Existing benchmarks evaluate LLMs and agents on specific tasks using answer accuracy \citep{cobbe2021gsm8k}, issue resolution rates \citep{jimenez2024swebench}, and task success rates \citep{zhou2024webarena}. Evaluation of multi-agent systems also covers collaboration and competition \citep{zhu2025multiagentbench}, coordination \citep{agashe2025coordination}, and collective reasoning \citep{du2024debate}. Together, these studies establish how LLM-based agents collaborate across tasks and how their individual and collective capabilities are evaluated.

\paragraph{Collective Intelligence and Structured Information Exchange.}
Group discussion allows members to exchange information and ideas that inform collective judgments \citep{stasser1985pooling,paulus2000brainwriting}. Established methods for structuring group discussion include the Nominal Group Technique \citep{delbecq1971group}, Delphi \citep{dalkey1963delphi}, Policy Delphi \citep{turoff1970policydelphi}, and Brainwriting \citep{paulus2000brainwriting}. Recent studies employ LLMs to simulate information propagation in agent societies \citep{liu2025fuse} and support group ideation \citep{shaer2024brainwriting}. Evaluation also examines the interaction process, including truth deviation during information propagation \citep{liu2025fuse}. Our work examines information flow in multi-agent systems by measuring how information becomes shared, expands beyond the initial source, and is retained or lost throughout interaction.
\section{\ours{}: Claim-level Information Dynamics Evaluation}
\label{sec: methodology}

\ours{} first decomposes agent statements into atomic claims,
compares corresponding claim pairs across agents and stages, and then aggregates these relations to evaluate them from three dimensions: \textbf{convergence}, \textbf{exploration}, and \textbf{retention}.
\textbf{Convergence} is measured by Informational Unity (U).
\textbf{Exploration} is measured by Unique Claim Count (CC)
and Extension Rate (ER).
\textbf{Retention} is measured by Original News Retention (OR),
Survival Rate (SR), and Claim Survival Length (SL). 
Algorithm~\ref{alg:information_evaluation} in \S\ref{app:pipeline} presents the pipeline of \ours{}.

\subsection{Preliminary}
We define several operators to support the evaluations in \ours{}.

\paragraph{Claim Decomposition.} A multi-agent interaction with $N$ agents is denoted as $\mathcal{A}
=\{a_1,a_2,\ldots,a_N\}$, where it employs a specific interaction strategy $\pi(\cdot)$. At the $t$-th interaction stage:
\begin{align}
    \boldsymbol{x^{(t)}}=\{x_i^{(t)}\}_{i=1}^N=\pi(\boldsymbol{x}^{(t-1)}),
\end{align}
where $x_i^{(t)}$ denotes the
public output produced by agent $a_i$ at stage $t$ and $x_i^{(0)}$ denotes the initial state. 

Given $x_i^{(t)}$, we decompose it through the atomic decomposition function $\mathcal{D}(\cdot)$:

\begin{align}
    C_i^{(t)}=\mathcal{D}\!\left(x_i^{(t)}\right)=
\left\{
c_{i,1}^{(t)},
c_{i,2}^{(t)},
\ldots,
c_{i,n_i^{(t)}}^{(t)}
\right\},
\end{align}
where $n_i^{(t)}=|C_i^{(t)}|$ is the number of extracted claims,
and $c_{i,j}^{(t)}$ is the $j$-th claim. Inspired by \citet{min2023factscore}, we employ LLMs as the decomposition function (see \S\ref{app:decomposition}). It ensures that each atomic claim expresses
one piece of information, which provides a common unit for comparing what agents express at a given stage and tracking what remains represented at later stages.

\paragraph{Semantic correspondence.} We measure semantic relevance between claims to identify the corresponding propositions.
For two
claims $c$ and $d$, their semantic correspondence score is assigned by an LLM (see \S\ref{app:semantic_judgments}) $R(c,d)\in\{0.0, 0.1, \ldots, 1.0\}$. A higher $R(c,d)$ indicates stronger correspondence at the level of
a specific proposition. To distinguish specific informational correspondence from broad topical similarity, we apply a hard-thresholding operation to define the effective semantic correspondence score as
\begin{align}
    \tilde{R}(c,d)=R(c,d)\mathbb{I}[R(c,d)\geq \tau],
\end{align}
where $\mathbb{I}[\cdot]$ denotes the indicator function, and we set $\tau$ as $0.75$ in practice. This formulation retains the original correspondence strength for claim pairs satisfying the threshold criterion while assigning zero to pairs below the threshold.

\paragraph{Claim Partitioning.} This operator is proposed to determine whether each agent claim $c_{i,j}^{(t)}\in C_i^{(t)}$ is supported by the source or introduces information beyond it. Given a source text $\boldsymbol{S}$, $S_0=\mathcal{D}(\boldsymbol{S})=\{s_1,s_2,\ldots,s_M\}$ denotes the $M$ atomic information units extracted from the complete initial text.

We employ an LLM-as-a-judge (see \S\ref{app:semantic_judgments}) to calculate the pairwise supporting score between a source claim $s\in S_0$ and an agent claim $c\in C_i^{(t)}$: $R_{\text{src}}(s,c) \in \{0.0,0.1,\ldots,1.0\}$, where a higher score indicates stronger support from the source unit
$s$ for the atomic agent claim $c$. After that, we apply a max-pooling operation to obtain the support score of claim $c$ from the source text:
\begin{align}
R_{\text{src}}^{\star}(c)=\max_{s\in S_0}R_{\text{src}}(s,c).
\end{align}

Finally, we employ a threshold $\tau_{\text{src}}$ to determine the support and extend sets:
\begin{align}
    S_i^{(t)}=\left\{c\in C_i^{(t)}\mid R_{\text{src}}^{\star}(c)\geq\tau_{\text{src}}\right\},\\
    E_i^{(t)}=\left\{c\in C_i^{(t)}\mid R_{\mathrm{src}}^{\star}(c)<\tau_{\text{src}}\right\},
\end{align}
where we set $\tau_{\text{src}}$ as $0.75$ in practice. In summary, a claim is source-supported when it repeats information already contained in the source. A claim is treated as an extension when it introduces additional information, such as a new fact or cause. Extension therefore denotes information that goes beyond the initial source and broadens the scope of the discussion.

\subsection{Convergence Dimension}
\label{sec:convergence}

Agents achieve convergence when they express the same information
and share similar views on it.

\noindent\textbf{Informational Unity (U)} measures how strongly agents converge on information and aligned viewpoints. A higher Informational Unity indicates that agents express more mutually corresponding specific information and take more similar viewpoints toward that information. A lower Unity can arise either because agents express different information or because they take different positions on the same information.
At stage $t$:
\begin{align}
    U^{(t)}=\frac{2}{N(N-1)}\sum_{i<j}U_{a_i a_j}^{(t)},
\end{align}
where $U_{a_i a_j}^{(t)}$ denote the score between an unordered agent pair $(a_i, a_j)$:
\begin{align}
    U_{a_i a_j}^{(t)}=\frac{U_{a_i\rightarrow a_j}^{(t)}+U_{a_j\rightarrow a_i}^{(t)}}{2},
\end{align}
where $U_{a_i\rightarrow a_j}^{(t)}$ denote the directed score. If an information unit from $a_i$ has no corresponding information in $a_j$, its contribution in the $a_i\rightarrow a_j$ direction is zero. Directed matching is necessary because the information sets expressed by two agents need not be symmetric: one agent may contain information not mentioned by the other. We calculate it by:
\begin{align}
    U_{a_i\rightarrow a_j}^{(t)}=\frac{1}{|C_i^{(t)}|}\sum_{c\in C_i^{(t)}}\max_{d\in C_j^{(t)}}m(c,d),
\end{align}
where $m(c,d)=\tilde{R}(c,d)A(c, d)$ is the claim alignment score. Following the formulation of opinion similarity by \citet{betz2022opinion}, we calculate $A(c,d)$ as 
\begin{align}
    A(c,d) = 1 - |\sigma(c)-\sigma(d)|,
\end{align}
where $A(c,d)=1$ indicates identical stance scores, and the value decreases as their stance scores diverge. Meanwhile, $\sigma(\cdot)$ denotes the stance score, which is given directly by LLMs (see \S\ref{app:decomposition}).

\subsection{Exploration Dimension}
\label{sec:exploration}
Exploration characterizes the range of information which agents explore during interaction.

\noindent\textbf{Unique Claim Count (CC)} measures how many distinct claims the group expresses after semantic grouping. 
A higher CC indicates that the group expresses a wider range of distinct information. A lower CC indicates that its claims are concentrated around fewer distinct pieces of information.
At stage $t$,
\begin{align}
    CC^{(t)}=\left|\mathcal{K}_{\text{src}}^{(t)}\right|+\left|\mathcal{K}_{\text{ext}}^{(t)}\right|,
\end{align}
where $\mathcal{K}_{\text{src}}^{(t)}$ and
$\mathcal{K}_{\text{ext}}^{(t)}$ denote the resulting cluster sets of $\mathcal{S}^{(t)}$ and $\mathcal{E}^{(t)}$:
\begin{align}
    \mathcal{S}^{(t)}=\bigcup_{i=1}^{N} S_i^{(t)},\qquad\mathcal{E}^{(t)}=\bigcup_{i=1}^{N} E_i^{(t)}.
\end{align}

We cluster source-supported claims and extension claims separately using a greedy algorithm. Claims with more corresponding claims from other agents are prioritized for processing. A claim can join a cluster only if its effective semantic correspondence with every member is positive and no claim in this cluster comes from the same agent. Among eligible clusters, larger clusters are preferentially chosen, followed by those with higher minimum correspondence scores with the new claim, then higher mean scores. If no cluster is eligible, a new cluster containing only this claim is created.

\noindent\textbf{Extension Rate (ER)} measures the proportion of expressed claims that extend beyond the initial source. A higher ER indicates that a larger proportion of the current output contains more additional information beyond the initial source, whereas a lower ER indicates that the discussion remains more strongly anchored to source-supported information.
At stage $t$:
\begin{align}
    ER^{(t)}=\frac{1}{N}\sum_{i=1}^{N}ER_i^{(t)}=\frac{1}{N}\sum_{i=1}^{N}\frac{|E_i^{(t)}|}{|C_i^{(t)}|}.
\end{align}

\subsection{Retention Dimension}
\label{sec:retention}
Information expressed early in an interaction may be preserved or absent from later statements.

\noindent\textbf{Original News Retention (OR)} measures how much initial source information remains represented in agents' statements. A higher OR indicates that more of the original information remains represented as the discussion develops, whereas a lower OR indicates that more source information has disappeared from agents' current outputs.
At stage $t$,
\begin{align}
    OR^{(t)}=\frac{1}{N}\sum_{i=1}^{N}OR_i^{(t)},
\end{align}
where $OR_i^{(t)}$ denotes the agent-level score:
\begin{align}
    OR_i^{(t)}=\frac{1}{|S_0|}\sum_{s\in S_0}\max_{c\in C_i^{(t)}}P_{\text{OR}}(s,c),
\end{align}
where $P_{\text{OR}}(s,c)\in\{0.0,0.2,0.5,0.8,1.0\}$ denotes the directional preservation score between a source unit $s\in S_0$ and a target claim $c\in C_i^{(t)}$. We apply an LLM to obtain this score (see \S\ref{app:semantic_judgments}).

\noindent\textbf{Survival Rate (SR)} measures how much extension information remains represented across two stages. An extension may appear in one stage and disappear in the next. We compare earlier extensions with later claims to distinguish retained content from transient content. A higher SR indicates that extension information is more strongly preserved in later information states, whereas a lower SR indicates that more extensions are transient.
At stage $t$, let $t'$ denote a later stage with $t'>t$. We calculate it by:
\begin{align}
    SR_{t\rightarrow t'}=\frac{\sum_{i=1}^{N}\sum_{e\in E_i^{(t)}}q_{t\rightarrow t'}(e)}{\sum_{i=1}^{N}|E_i^{(t)}|},
\end{align}
where $q_{t\rightarrow t'}(e)$ denotes the preservation of a source-stage extension $e$:
\begin{align}
    q_{t\rightarrow t'}(e)=\max_{e'\in E_i^{(t')}}P_{\text{SR}}(e,e'),
\end{align}
where $P_{\text{SR}}(e,e')\in\{0.0,0.2,0.5,0.8,1.0\}$ is the directional preservation score obtained by the LLMs (see \S\ref{app:semantic_judgments}).

\noindent\textbf{Claim Survival Length (SL)} measures how long extension claims remain traceable. A higher $SL(d)$ indicates that a larger proportion of information sets survives at least $d$ positions.
We measure it through the proportion of information sets surviving at least each possible duration:
\begin{align}
    SL(d)&=\frac{\overline{W}(d)}{\overline{W}(1)},\qquad d=1,\ldots,T,\\
    \overline{W}(d)&=\frac{1}{|\mathcal{H}|}\sum_{u\in\mathcal{H}}W_u(d),
\end{align}
where $\mathcal{H}$ is the set of evaluated interaction units, $W_u(d)$ counts the information sets surviving at least $d$ positions in unit $u$, and $T$ is the maximum number of observable positions, 
\begin{align}
    W_u(d)=\sum_{q\in\mathcal Q_{u}}\mathbb I[S(q)\geq d],
\end{align}
where $\mathcal Q_{u}$ contains all constructed information sets in unit $u$. The survival length $S(q)$ is calculated as
\begin{align}
    S(q)=l_q-r_q+1,
\end{align}
where $l_q$ and $r_q$ denote its last and first position indices, respectively.

We construct information sets $q$ using extension claim instances as nodes. Claims from different agents at the same position are connected in both directions when $R(c,d)\geq\tau$. Between adjacent positions, an earlier claim is connected to a later claim along the specified propagation path when $P_{\mathrm{SR}}(c,d)\geq\tau$.

We process claims in position order. Each unassigned claim starts a new information set $q$ containing all reachable, unassigned claims. Each claim instance belongs to only one set, and new sets may originate at any position.

\subsection{Validation}
We externally validate the convergence dimension because, compared with the other two dimensions, it operationalizes a higher-level notion of information convergence by combining cross-agent semantic correspondence with stance alignment.

\noindent\textbf{The convergence of \ours{} is aligned with humans.} We apply three human evaluation datasets, including DBS-eval \citep{tauchmann2020dbseval}, DELPHISTAR \citep{niederberger2024delphi}, and Polis \citep{small2021polis} (see \S\ref{app:human_validation}). These datasets provide reference scores of information sharing or agreement based on participants' expressions. We compute the Pearson and Spearman correlations between \ours{}'s Informational Unity and the reference scores of these datasets, as shown in Table~\ref{tab:unity_validation}. The results support its alignment with human assessments of information agreement.

\noindent\textbf{\ours{} can better capture the convergence than direct prompting.} We directly prompt LLMs to generate the convergence scores as the Simple baseline and present the results of DBS-eval in Table~\ref{tab:unity_simple}. This consistent advantage supports the value of claim decomposition and alignment for evaluating shared information.

\noindent\textbf{\ours{} is robust across different LLM evaluators.} We further apply two more LLMs as evaluators to evaluate DBS-eval and report the correlations in Table~\ref{tab:evaluator_consistency}. Both linear and rank correlations indicate consistency in how the evaluators score these topics.

\begin{table}[t]
\centering
\small
\begin{tabular}{lrrr}
\toprule
\textbf{Dataset} & \textbf{$n$} & \textbf{Pearson $r$} & \textbf{Spearman $\rho$} \\
\midrule
DBS-eval & 10 & 0.641 & 0.638 \\
DELPHISTAR & 33 & 0.729 & 0.689 \\
Polis & 16 & 0.994 & 0.840 \\
\bottomrule
\end{tabular}
\caption{\textbf{Correlations between Informational Unity and the scores from three human evaluation datasets.} \ours{} is aligned with humans.}
\label{tab:unity_validation}
\end{table}

\begin{table}[t]
\centering
\small
\begin{tabular}{lrrr}
\toprule
\textbf{Model} & \textbf{Unity} & \textbf{Simple} & \textbf{$\Delta r$} \\
\midrule
Gemini 2.5 Flash Lite & \textbf{0.641} & -0.110 & +0.751 \\
Gemini 2.5 Flash & \textbf{0.588} & -0.031 & +0.619 \\
Gemini 3.8 Flash & \textbf{0.678} & 0.620 & +0.058 \\
\bottomrule
\end{tabular}
\caption{\textbf{Correlations with human reference ratings on DBS-eval compared to the Simple baseline.} $\Delta r=r_{\mathrm{Unity}}-r_{\mathrm{Simple}}$. \ours{} performs better than the Simple baseline, namely, direct prompting.}
\label{tab:unity_simple}
\end{table}

\begin{table}[t]
\centering
\small
\begin{tabular}{lrr}
\toprule
\textbf{Evaluator pair} & \textbf{Pearson $r$} & \textbf{Spearman $\rho$} \\
\midrule
2.5 Flash Lite / 2.5 Flash & 0.971 & 0.939 \\
3.8 Flash / 2.5 Flash Lite & 0.765 & 0.770 \\
3.8 Flash / 2.5 Flash & 0.807 & 0.842 \\
\bottomrule
\end{tabular}
\caption{\textbf{Correlations of Informational Unity scores across different LLM evaluators.} \ours{} is robust under different LLMs as evaluators.}
\label{tab:evaluator_consistency}
\end{table}

\section{Experimental Settings}
\paragraph{Discussion Setting.} We evaluate information dynamics in controlled multi-agent news discussions. We use 40 short news cases derived from 20 Reuters reports,
with one Event-only case and one Event+Seed case per report.
The source reports and their original URLs are provided in
\S\ref{app:news_sources}. For each case, we construct a group of five agents with fixed personas and run the same case under five interaction strategies: \textbf{Free}, \textbf{NGT}~\citep{delbecq1971group}, \textbf{Delphi}~\citep{dalkey1963delphi}, \textbf{Policy Delphi}~\citep{turoff1970policydelphi}, and \textbf{Brainwriting}~\citep{paulus2000brainwriting}. NGT, Delphi, and Policy Delphi employ repeatedly pooled structures, whereas Brainwriting uses a distributed relay structure. Table~\ref{tab:interaction_methods} of \S\ref{app:interaction_strategies} summarizes their
interaction structures and method-specific constraints. We apply \texttt{GPT-4o-mini} as the LLM for discussion in the experiments. Each case contains four independent interaction cycles, with four stages in each cycle, followed by a final stage in which agents privately summarize their previous outputs. Each cycle starts independently from the same source news (details in \S\ref{app:discussion_schedule}).
\begin{table*}[t]
\centering
\small
\begin{tabular}{llcccccc}
\toprule[1.5pt]
\multirow{2}{*}{\textbf{Strategies}} & \multirow{2}{*}{\textbf{Setting}} & \textbf{Convergence} & \multicolumn{2}{c}{\textbf{Exploration}} & \multicolumn{3}{c}{\textbf{Retention}} \\
\cmidrule[0.5pt](lr){3-3}\cmidrule[0.5pt](lr){4-5}\cmidrule[0.5pt](lr){6-8}
&& \textbf{U} & \textbf{CC} & \textbf{ER} & \textbf{OR} & \textbf{SR} & \textbf{SL} \\
\midrule[1pt]
\textbf{Free} & Control & 0.371 & 13.46 & 0.892 & 0.713 & 0.680 & 1.521 \\
\midrule[1pt]
\multirow{2}{*}{\textbf{NGT}} & Full & 0.323 & 13.24 & 0.943 & 0.513 & 0.643 & 1.391 \\
& Framework & 0.518\textsubscript{+60.6\%} & 10.41\textsubscript{-21.4\%} & 0.889\textsubscript{-5.7\%} & 0.766\textsubscript{+49.3\%} & 0.721\textsubscript{+12.1\%} & 1.616\textsubscript{+16.2\%} \\
\midrule[1pt]
\multirow{2}{*}{\textbf{Delphi}} & Full & 0.275 & 14.92 & 0.920 & 0.630 & 0.635 & 1.419 \\
& Framework & 0.497\textsubscript{+80.4\%} & 10.88\textsubscript{-27.1\%} & 0.887\textsubscript{-3.6\%} & 0.750\textsubscript{+19.0\%} & 0.703\textsubscript{+10.8\%} & 1.544\textsubscript{+8.8\%} \\
\midrule[1pt]
\multirow{2}{*}{\textbf{Policy Delphi}} & Full & 0.244 & 16.53 & 0.943 & 0.546 & 0.620 & 1.435 \\
& Framework & 0.486\textsubscript{+99.2\%} & 10.62\textsubscript{-35.8\%} & 0.933\textsubscript{-1.1\%} & 0.682\textsubscript{+24.9\%} & 0.708\textsubscript{+14.2\%} & 1.508\textsubscript{+5.1\%} \\
\midrule[1pt]
\multirow{2}{*}{\textbf{Brainwriting}} & Full & 0.172 & 18.18 & 0.902 & 0.693 & 0.668 & 1.656 \\
& Framework & 0.302\textsubscript{+75.0\%} & 14.31\textsubscript{-21.3\%} & 0.888\textsubscript{-1.5\%} & 0.746\textsubscript{+7.6\%} & 0.753\textsubscript{+12.6\%} & 1.977\textsubscript{+19.4\%} \\
\bottomrule[1.5pt]
\end{tabular}
\caption{\textbf{Aggregate results across 40 news cases across five interaction strategies.} Metrics are grouped into three dimensions: convergence (U), exploration (CC, ER), and retention (OR, SR, SL). Subscripts report percentage changes from Full to Framework,
with Full as the baseline.}
\label{tab:overall_results}
\end{table*}

\paragraph{Interaction Conditions.}
Free strategy serves as the control condition. It allows agents to discuss without a prescribed interaction procedure or behavioral constraints. The other four strategies are based on classical group interaction methods, with adjustments for multi-agent systems. For each of the four classical interaction strategies, the Full setting retains the complete method, while the Framework setting removes the additional constraints and preserves the core interaction structure.

\paragraph{Evaluation Metrics.} We consider each news case as an independent statistical unit. We apply the \texttt{gemini-2.5-flash-lite} as the main evaluation LLM (details in \S\ref{app:implementation}). The main text reports aggregate metric values, and computation and aggregation rules are provided in \S\ref{app:metric_details}. We also provide complete metric trajectories in \S\ref{app:additional_results}.

\section{Results}
\begin{figure*}[t]
\centering
\includegraphics[width=\textwidth]{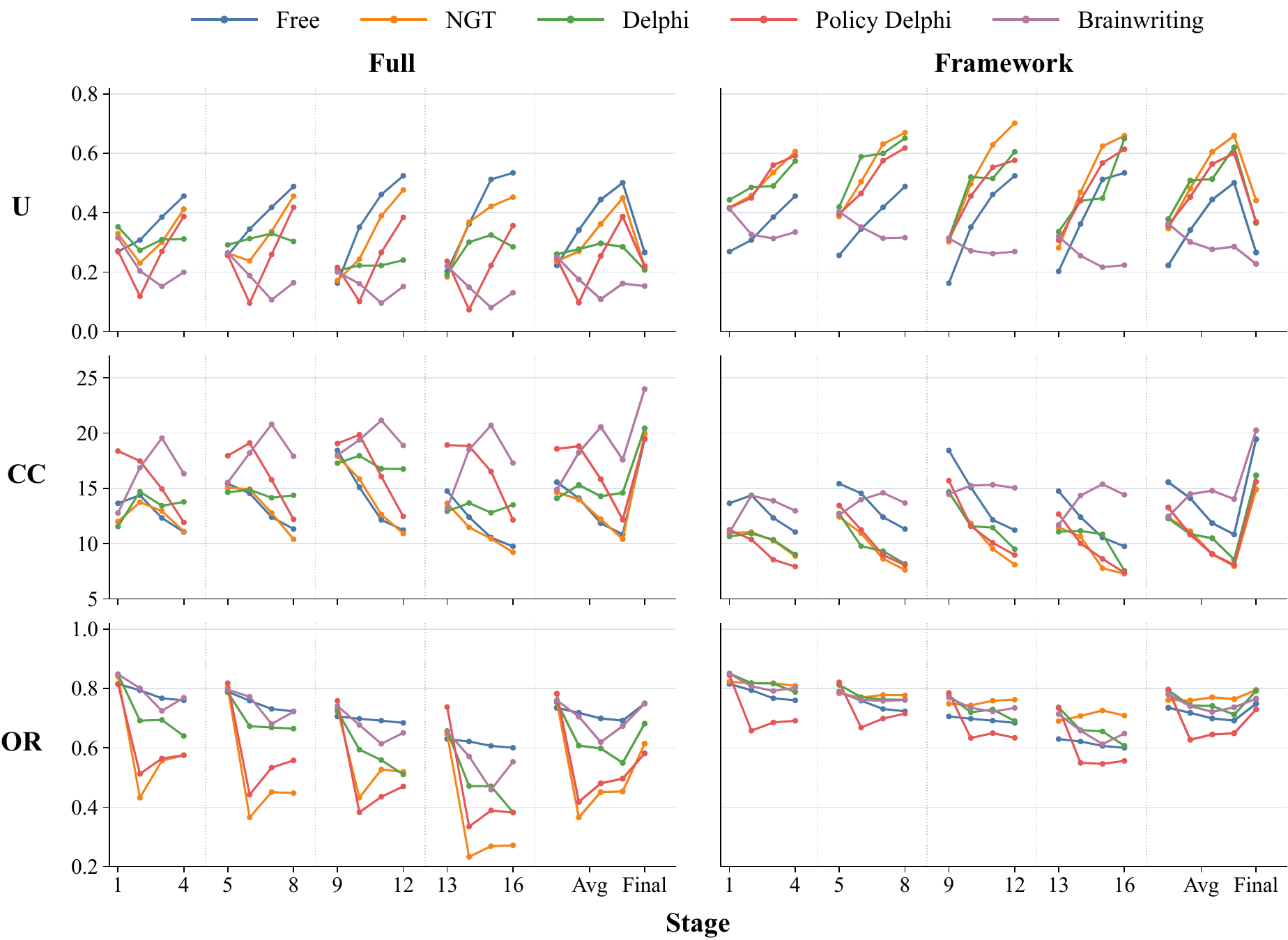}
\caption{\textbf{Trajectories of U, CC, and OR across 40 news cases.} Avg shows the mean trajectory of each cycle. Final denotes Stage 17. \ours{} can capture the information dynamics during the multi-agent discussion processes.}
\label{fig:metric_overview}
\end{figure*}

Table~\ref{tab:overall_results} reports aggregate metrics, while Figure~\ref{fig:metric_overview} shows the trajectories of three representative metrics: U, CC, and OR. Trajectories for ER, SR, and SL are provided in \S\ref{app:additional_results}. We can conclude that:

\noindent\textbf{Interaction structures and constraints shape information dynamics.}
Across interaction strategies, repeated information pooling leads to greater information sharing and a narrower information space, whereas distributed relay maintains broader but less shared information. Comparisons between Full and Framework settings show that additional constraints increase exploration while weakening convergence and retention across all four structured methods.

\noindent\textbf{Pooling and relay generally produce different convergence trajectories.}
Under Framework settings, the three structured pooling methods and Brainwriting have nearly equal U after independent generation (0.361 on average vs.\ 0.362), but diverge by the Cycle-end synthesis or revision stage (0.626 on average vs.\ 0.286). U increases in only 28 of 160 case--Cycle observations for Brainwriting Full and 33 for Brainwriting Framework, both less than the 90 for Delphi Full and 135--156 for the other pooling conditions.

\noindent\textbf{Exploration can be influenced by both interaction structures and additional constraints.}
As with convergence, exploration follows different trajectories across interaction structures. Under Framework settings, CC gradually decreases in the three structured pooling methods. On average across these methods, successive stage-to-stage reductions in CC are 13.5\%, 11.0\%, and 10.5\% of Stage-1 CC. By contrast, Brainwriting's stage-to-stage changes are +16.4\%, +2.5\%, and $-6.1\%$ of Stage-1 CC, showing an initial increase followed by smaller changes. Moving from Full to Framework reduces aggregate CC by 26.5\% across the four strategies, while ER decreases by 3.0\% on average. The reduction in CC occurs in all 40 paired cases for each strategy, indicating that removing additional constraints narrows the claim space.

\noindent\textbf{Information retention varies across interaction structures and improves when additional constraints are removed.} Under Full settings, OR decreases from Stage 1 to Stage 4 by 34.6\% on average across the three pooling methods, exceeding the 11.4\% decline in Brainwriting. Meanwhile, SR rises within Cycles across the pooling methods, while decreasing in Brainwriting. Moving from Full to Framework improves long-term retention more than short-term retention. On average across the four strategies, the proportion of extension lineages surviving at least two positions increases by only 0.9 percentage points. The increase is 6.0 percentage points for lineages surviving at least three positions. Interestingly, we observe that information omitted by some agents can reappear after subsequent sharing, with mean OR recovering by 0.019 from Stage 2 to Stage 4 under Full settings.

\noindent\textbf{Greater exploration tends to come at the expense of convergence and retention.}
Convergence and exploration show opposing patterns, with mean Pearson and Spearman coefficients of $-0.815$ and $-0.825$ for U--CC across the nine conditions. Exploration and retention also show opposing patterns, with a mean Pearson coefficient of $-0.710$ for CC--SR and a mean Spearman coefficient of $-0.666$ for ER--OR. Convergence and extension retention are positively associated across the seven non-Brainwriting conditions, with mean Pearson and Spearman coefficients of 0.798 and 0.787 for U--SR. The corresponding coefficients in Brainwriting are 0.443 and 0.317 under Full settings, and 0.402 and 0.027 under Framework settings. Its distributed structure allows information to persist locally without becoming widely shared.
\section{Discussion}

Group discussion supports information exchange and collective judgments \citep{stasser1985pooling,paulus2000brainwriting}. Interaction structures determine what information agents can access. Different interaction structures favor different patterns of information propagation. Free permits flexible interaction, but access to all previous responses still gives rise to pooling. Repeated pooling promotes convergence through repeated sharing, while distributed relay encourages exploration along separate paths. Its longer claim lineages show that information can persist along a relay path without becoming widely shared. Behavioral constraints shape content selection. Differentiated views or new details may displace shared and source information in limited output space, increasing exploration while weakening convergence and retention.

Compared with previous work, our study places greater emphasis on information dynamics among agents and shows how information is shared, extended, and retained at the level of individual claims. We observe that source information can reappear and that different strategies favor different patterns of information sharing and persistence. These observations show that interaction settings affect both what agents express and how information spreads. Our framework helps researchers examine information dynamics and understand how interaction settings organize information within groups.

\section{Conclusion}
\label{sec:conclusion}


We propose \ours{}, a claim-level framework for analyzing multi-agent information flow through \textbf{convergence}, \textbf{exploration}, and \textbf{retention}. Across five interaction strategies, pooled interaction promotes convergence but narrows the information space, while distributed relay preserves broader, less shared information. Comparisons between Framework and Full show that additional behavioral constraints increase exploration while weakening convergence and retention. These results suggest that multi-agent behavior is better understood through underlying information-flow mechanisms than through final outcomes.

\section*{Limitations}

Our experiments use one generation model and five fixed personas across all interaction conditions. This design supports controlled comparisons, but does not examine how model capabilities or different persona compositions influence information dynamics. Further experiments with other models and agent profiles would help assess the generality of the observed patterns.

Each Cycle starts independently from the same news input, without carrying discussion history or private memory between Cycles. This allows comparisons across discussion objectives, but does not capture information accumulation across sustained interactions. Future work could examine longer discussions in which agents retain and revise information across multiple Cycles.
\section*{Ethical Considerations}

Our experiments use publicly reported news and synthetic agent profiles. The source reports are attributed in \S\ref{app:news_sources}, and experimental inputs are concise paraphrases rather than full reproductions of the articles. Generated discussions may contain unsupported explanations, predictions, or proposals and should not be treated as verified reporting. The synthetic profiles are not intended to represent identifiable individuals.

Human validation reuses existing datasets without recruiting new participants. These data are used to evaluate information sharing and agreement rather than to assess individual participants. The framework measures expressed information. High convergence does not imply factual correctness, and low convergence does not imply undesirable disagreement.
\bibliography{custom}

\appendix

\section{Implementation Details}
\label{app:implementation}

\subsection{Discussion Schedule}
\label{app:discussion_schedule}
Each of the 40 news cases is discussed by five agents with fixed personas. Discussions are generated using \texttt{gpt-4o-mini}. The same news cases, personas, cycle objectives and stage schedule are used across five interaction strategies and their applicable Full and Framework settings.

Each case contains four independent Cycles of four Stages each, followed by Stage 17 for private summaries. Stage 1 of each Cycle starts from the same original news, the Cycle objective, and each agent's fixed persona. Interaction proceeds through Stages 1--4 within a Cycle. Neither public outputs nor private states are passed between Cycles. At Stage 17, each agent privately integrates its own final states from the four Cycles into a final summary, without further group exchange.
\begin{table}[t]
\centering\small
\begin{tabular}{clp{0.58\columnwidth}}
\toprule
\textbf{Cycle} & \textbf{Stages} & \textbf{Discussion focus} \\
\midrule
1 & 1--4 & Possible causes and problem framing \\
\midrule
2 & 5--8 & Direct and indirect consequences \\
\midrule
3 & 9--12 & Stakeholder reactions and risks \\
\midrule
4 & 13--16 & Responses or policy options \\
\bottomrule
\end{tabular}
\caption{Cycle objectives and global Stage indices. Stage 17 is the private summary stage.}
\label{tab:cycle_schedule}
\end{table}
The five agent profiles in Table~\ref{tab:agent_personas} are fixed across news cases, interaction methods, and experimental settings, and are inserted into the generation prompt.

\begin{table*}[t]
\centering\small
\begin{tabular}{cllp{0.30\textwidth}l}
\toprule
\textbf{Agent ID} & \textbf{Name} & \textbf{Role} &
\textbf{Persona} & \textbf{Qualification} \\
\midrule
0 & James & Commentator & analytical, outspoken, cooperative & Bachelor's Degree \\
1 & Mary & Verifier & precise, skeptical, patient & Master's Degree \\
2 & Robert & Spreader & gregarious, quick, optimistic & High School Diploma \\
3 & Patricia & Bystander & reserved, empathetic, cautious & Associate Degree \\
4 & John & Commentator & independent, insightful, stubborn & PhD \\
\bottomrule
\end{tabular}
\caption{Fixed agent profiles used across all experimental conditions.}
\label{tab:agent_personas}
\end{table*}
\subsection{Interaction Strategies}
\label{app:interaction_strategies}
Free serves as the control condition. For NGT, Delphi, Policy Delphi, and Brainwriting, Full retains the complete method; Framework removes the additional constraints while preserving the core interaction structure. Table~\ref{tab:interaction_methods} summarizes these structures and constraints.\begin{table*}[t]
\centering\small
\begin{tabular}{p{0.13\textwidth}p{0.39\textwidth}p{0.39\textwidth}}
\toprule
\textbf{Strategy} & \textbf{Core interaction structure} & \textbf{Additional constraints in Full} \\
\midrule
Free & Open discussion without a prescribed interaction procedure & None \\
\midrule
NGT & Independent generation; anonymous sharing; clarification; private synthesis & Equal extension sharing; protection of low-frequency views \\
\midrule
Delphi & Anonymous feedback; private multi-round revision & Coverage; minority views; pro/con reasons; and uncertainty requirements \\
\midrule
Policy Delphi & Anonymous presentation of stances and reasons; responses; private disagreement summary & Pro/con; alternatives; strongest opposing views; and self-criticism \\
\midrule
Brainwriting & Anonymous relay; distributed lineage; private lineage review & Risk; counterfactual; consequence; and causal-chain elaboration \\
\bottomrule
\end{tabular}
\caption{Interaction structures and additional constraints. Framework settings preserve the information-exchange arrangements while reducing or removing the requirements in the final column. In both NGT settings, agents generate ideas at Stage 1 and conduct subsequent discussion based on the existing idea bank.}
\label{tab:interaction_methods}
\end{table*}

\subsection{Atomic Decomposition and Stance Scoring}
\label{app:decomposition}
We evaluate agents' public outputs at each Stage and their private final summaries at Stage 17. Private thought and hidden reasoning are excluded. We use \texttt{gemini-3.1-flash-lite} to decompose both the original news and agent outputs into atomic claims and assign claim-level stance scores.

Each claim expresses one proposition. Decomposition preserves the actor, action, object, attribution, cause, consequence, condition, time, location, quantity, negation, and uncertainty when these affect meaning. For example, a claim that an organization states a proposition retains that attribution after decomposition.

The prompt requests a continuous stance score in $[0,1]$ for each complete claim, with three anchors: 0 indicates rejection, 0.5 indicates a neutral or uncertain stance, and 1 indicates strong support. Intermediate values are permitted. The implementation validates that the score lies within $[0,1]$. Scores reflect commitment to the whole canonical proposition toward an actor. An assertion that ``Officials said $P$'' supports the reporting proposition but does not by itself endorse $P$. 

\subsection{Prompt Templates}
\label{app:compact_prompts}
The prompt templates and output schemas below are simplified for presentation.
All methods use the shared generation template with the Stage-specific instructions in Table~\ref{tab:compact_stage_guidance}. Delphi and Policy Delphi use separate moderator prompts to compile anonymous feedback between Stages.
Braced placeholders are populated with the agent profile, Cycle objective, permitted material, or claims being evaluated. In the generation template, source contains the original news at the first Stage of each Cycle; at subsequent Stages, it is replaced with ``The source was read at position 1 and is not repeated here.'' Method-specific instructions are listed in Table~\ref{tab:compact_stage_guidance}.

\paragraph{Shared Generation.}
\begin{Verbatim}[fontsize=\small,breaklines=true,breakanywhere=true]
You are Agent {profile.agent_id} ({profile.name}) in a controlled multi-agent news experiment.
Method: {method}; cycle: {cycle}; position: {position}/4
Objective: {label}; question: {question}
Profile: {profile.role}; {profile.persona}; {profile.qualification}

Original news: {source}
Private memory: {private_memory}
Method-permitted material: {visible}
Instruction: {instruction}

Write a standalone public paragraph and a private memory summary. Distinguish
source facts from uncertain explanations, predictions, risks, or proposals.

Return one JSON object with all these fields:
{
  "news": "standalone public paragraph",
  "thought": "private memory summary",
  "action": "originate | elaborate | support | challenge | revise | synthesize | withdraw | repeat"
}
\end{Verbatim}

\paragraph{Private Final Summary.}
\begin{Verbatim}[fontsize=\scriptsize,breaklines=true,breakanywhere=true]
You are Agent {profile.agent_id} ({profile.name}). Produce a private final summary.
Read only your own position-4 state from each cycle. Integrate the four
objectives without starting a new discussion; preserve uncertainty.
Method: {method}
Own cycle-end states: {cycle_ends}
Return one JSON object with all these fields:
{
  "news": "standalone final summary",
  "thought": "private memory summary",
  "action": "synthesize"
}
\end{Verbatim}
\paragraph{Moderator Feedback.}
For Delphi and Policy Delphi, moderator inputs contain anonymous participant IDs, public outputs, and extension items. The shared template below uses the setting-specific instructions that follow.

\begin{Verbatim}[fontsize=\scriptsize,breaklines=true,breakanywhere=true]
You are the neutral moderator for cycle {cycle}, stage {stage}.
{instruction}

Inputs:
{packet}

{output_instruction}
\end{Verbatim}

\paragraph{Delphi Full.}
\begin{Verbatim}[fontsize=\scriptsize,breaklines=true,breakanywhere=true]
Create anonymous controlled feedback: canonical details, coverage counts,
supporting and opposing stances, representative reasons, low-frequency views,
and unresolved uncertainties. Do not identify authors or call the majority correct.

Return JSON only: {"feedback": "complete moderator packet"}
\end{Verbatim}

\paragraph{Policy Delphi Full.}
\begin{Verbatim}[fontsize=\scriptsize,breaklines=true,breakanywhere=true]
Create an anonymous balanced pro-con matrix with canonical detail, pro reasons,
con reasons, alternatives, uncertainty, and coverage. Preserve minority views.

Return JSON only: {"feedback": "complete moderator packet"}
\end{Verbatim}

\paragraph{Delphi Framework.}
\begin{Verbatim}[fontsize=\scriptsize,breaklines=true,breakanywhere=true]
Create a faithful anonymous compilation of the supplied material. Preserve all
distinct substantive items and lightly merge only near-duplicates. Do not evaluate,
rank, endorse, balance, or seek consensus. Do not introduce any actor, fact, cause,
consequence, intervention, number, risk, prediction, policy option, counterargument,
or interpretation absent from the inputs. Do not manufacture disagreement or label
a repeated view as correct.

Return JSON in this shape:
{"feedback": [{"item": "an actual substantive anonymous item derived from the inputs"}]}
The feedback array must contain at least one actual input-derived item. Never return
the words "complete moderator packet", a placeholder, or instructions about what to write.
\end{Verbatim}

\paragraph{Policy Delphi Framework.}
\begin{Verbatim}[fontsize=\scriptsize,breaklines=true,breakanywhere=true]
Organize only positions and reasons actually present into an anonymous issue/position
map. For each issue, use observed positions, support, opposition, alternatives, and
uncertainties only when supplied; any category may be empty. Do not invent missing
perspectives, counterarguments, alternatives, uncertainty, actors, facts, causes,
consequences, risks, predictions, or policy options. Do not balance categories
artificially, rank views, or treat repetition as correctness.

Return JSON in this shape:
{"feedback": [{"issue": "actual observed issue", "positions_observed": [],
"support_observed": [], "opposition_observed": [], "alternatives_observed": [],
"uncertainties_observed": []}]}
The feedback array must contain actual input-derived content. Empty category arrays are valid;
never invent content merely to fill them, and never return a placeholder.
\end{Verbatim}
\paragraph{Atomic Decomposition and Stance.}
\begin{Verbatim}[fontsize=\scriptsize,breaklines=true,breakanywhere=true]
Decompose the public output into self-contained atomic claims.
Preserve subject, action, object, cause, result, condition, time, place,
numbers, negation, uncertainty, modality, and source attribution. Evaluate only
the supplied public text. Assign a numerical stance_score in [0,1] to each
complete claim: 0 means rejection, 0.5 means neutral or uncertain, and 1 means
strong support.

PUBLIC OUTPUT:
{text}

Return one JSON object with this structure:
{
  "claims": [
    {
      "claim": "one atomic proposition",
      "stance_score": 0.5
    }
  ]
}
\end{Verbatim}

\paragraph{Source Support.}
\begin{Verbatim}[fontsize=\scriptsize,breaklines=true,breakanywhere=true]
Judge whether the complete TARGET CLAIM is directly supported
by the SOURCE UNIT. Score source support in {0.0, 0.1, ..., 1.0}. Added causes, consequences,
predictions, proposals, attributions, quantities, or other unsupported details
reduce support.

SOURCE UNIT:
{source_claim}

TARGET CLAIM:
{claim}

Return one JSON object with all these fields:
{
  "source_support": 0.0,
  "reason": "short explanation"
}
\end{Verbatim}

\paragraph{Semantic Correspondence.}
\begin{Verbatim}[fontsize=\scriptsize,breaklines=true,breakanywhere=true]
Score semantic correspondence R(c,d) in {0.0, 0.1, ..., 1.0}. Judge whether
the two atomic claims express the same substantive proposition. Ignore stance;
stance alignment is computed separately.

CLAIM A:
{claim_a}

CLAIM B:
{claim_b}

Return one JSON object with all these fields:
{
  "relevance": 0.0,
  "reason": "short slot-level explanation"
}
\end{Verbatim}

\paragraph{Directional Preservation.}
\begin{Verbatim}[fontsize=\scriptsize,breaklines=true,breakanywhere=true]
Judge directional preservation from the earlier claim to the later claim.
Use exactly one score from 0.0, 0.2, 0.5, 0.8, or 1.0.

EARLIER CLAIM:
{earlier}

LATER CLAIM:
{later}

Return one JSON object with all these fields:
{
  "preservation": 0.0,
  "reason": "short explanation"
}
\end{Verbatim}

\begin{table*}[t]
\centering
\scriptsize
\setlength{\tabcolsep}{3pt}
\begin{tabular}{
p{.10\textwidth}
p{.20\textwidth}
p{.20\textwidth}
p{.20\textwidth}
p{.20\textwidth}}
\toprule
\textbf{Setting} &
\textbf{Local Stage 1} &
\textbf{Local Stage 2} &
\textbf{Local Stage 3} &
\textbf{Local Stage 4} \\
\midrule
Free &
Start independently. &
Interact freely. &
Interact freely. &
Choose any useful action; consensus is optional. \\
\midrule
NGT Full &
Work silently and independently. &
Read the balanced idea bank; restate and clarify before judging. &
Clarify, support, challenge, revise, or combine shared claims. &
Privately synthesize retained, rejected, and uncertain claims. \\
NGT Framework &
Work silently and independently. Only this Stage permits new extension claims. &
Read the complete, balanced Stage-1 idea bank. Restate, compare, support, challenge, or clarify existing ideas without adding new claims. &
Discuss the Stage-1 idea bank and preceding sharing outputs. Select, support, challenge, clarify, revise wording, or combine existing claims without introducing new propositions. &
Privately synthesize only Stage-1 claims. Retain, reject, or express uncertainty without adding new propositions. Other Stage-4 outputs are hidden. \\
\midrule
Delphi Full &
Answer privately and independently. &
Use anonymous controlled feedback without treating frequency as truth. &
Revise privately. &
Use second feedback for a final private revision. \\
Delphi Framework &
Generate independently. &
Review the anonymous compilation and respond freely. &
Privately reconsider it. &
Use the second compilation for final revision. \\
\midrule
Policy Delphi Full &
Generate positions, pro/con reasons, alternatives, and uncertainty. &
Use the anonymous pro-con matrix and preserve minority positions. &
Develop counterarguments and alternatives. &
Map accepted, rejected, uncertain, and disputed claims. \\
Policy Delphi Framework &
Respond independently without fixed categories. &
Review the anonymous position map and choose any action. &
Respond privately to observed positions. &
Privately synthesize your position. \\
\midrule
Brainwriting Full &
Generate independently. &
Elaborate the assigned anonymous item. &
Elaborate the next assigned lineage item. &
Privately synthesize the lineage. \\
Brainwriting Framework &
Generate independently. &
Read one assigned anonymous item and respond freely. &
Read the next lineage item and respond freely. &
Review only the routed lineage and synthesize it. \\
\bottomrule
\end{tabular}
\caption{Abbreviated method-specific instructions used at each local Stage within a Cycle.}
\label{tab:compact_stage_guidance}
\end{table*}

\subsection{Runtime Configuration}
\label{app:runtime_configuration}
Agent generation and Stage-17 summaries use a temperature of 0.4. Delphi and Policy Delphi moderator calls use 0.1. Atomic decomposition, stance scoring, semantic evaluation, and human-validation evaluator calls use 0.0. We do not explicitly set other settings such as max tokens, leaving them at their API defaults. Only outputs that pass parsing and schema validation are stored in the evaluation cache.
\subsection{Semantic Judgments}
\label{app:semantic_judgments}
We use \texttt{gemini-2.5-flash-lite} for the independent source-support audit, semantic correspondence, and directional preservation judgments. 

Source support measures whether the source unit supports the complete agent claim. Semantic correspondence concerns a specific proposition and follows the proposition-slot constraint: an added substantive slot cannot be ignored when treating claims as corresponding. Source support and semantic correspondence are scored on the grid $\{0.0,0.1,\ldots,1.0\}$. Directional preservation is scored on $\{0.0,0.2,0.5,0.8,1.0\}$. Both use a threshold of 0.75, so the minimum passing score is 0.8.
\section{Metric Details and Robustness}
\label{app:metric_details}

\subsection{Evaluation Positions and Propagation Paths}
\label{app:metric_positions}
U, CC, ER, and OR are computed at each of the 17 Stages. SR and SL track extension claims only and are computed separately within each Cycle, with Stage 17 appended:
\[
(1,2,3,4,17),\quad(5,6,7,8,17),
\]
\[
(9,10,11,12,17),\quad(13,14,15,16,17).
\]
SR compares adjacent positions. For both SR and SL, cross-position comparisons follow the same agent, except in Brainwriting, where they follow the relay path across agents defined by the method. Each agent's final summary is reused independently in each Cycle. No temporal edges connect different Cycles. SL counts positions, and each claim instance belongs to only one information set within an interaction unit.

\subsection{Claim Clustering}
\label{app:claim_clustering}
Source-supported and extension claims are clustered separately. Claims are processed in descending order of the number of corresponding claims from other agents; ties use Agent ID and claim ID. A claim joins a cluster only when it meets the semantic correspondence threshold with every existing member and that cluster contains no claim from the same agent.

Among eligible clusters, larger clusters are chosen preferentially, followed by higher minimum correspondence with the new claim, then higher mean correspondence. Remaining ties favor the earliest-created cluster. If none is eligible, a new singleton cluster is created. Existing clusters are not merged. Clustering follows a greedy complete-link rule.

\subsection{Aggregate Values}
\label{app:aggregation}
U, CC, ER, OR, and SR are averaged equally across the 40 news cases; the five agents within a case are not independent statistical samples. For SL, $\mathcal{H}$ contains $40\times4=160$ case--Cycle units per condition. Counts $W_u(d)$ are averaged across these units and normalized as defined in the Method section. This aggregation is equivalent to pooling counts across units.

The aggregate values in Table~\ref{tab:overall_results} are
\begin{equation}
\overline{X}=\frac{1}{17}\sum_{t=1}^{17}X^{(t)},
\qquad X\in\{U,CC,ER,OR\},
\end{equation}
\begin{equation}
\overline{SR}=\frac{1}{16}\sum_{(t,t')\in\mathcal{T}}SR_{t\rightarrow t'},
\end{equation}
\begin{equation}
\overline{SL}=\sum_{d=1}^{5}SL(d).
\end{equation}
Here, $\mathcal{T}$ contains the 16 evaluated transitions, including each Cycle's final Stage to Stage 17. $\overline{SL}$ is the mean survival length of extension information sets under this aggregation.

\subsection{Evaluation Pipeline}
\label{app:pipeline}
Algorithm~\ref{alg:information_evaluation} summarizes the evaluation pipeline. Here, $u$ indexes a case--Cycle unit and $\mathcal{P}_u$ specifies its ordered positions. The local set $\mathcal{Q}_u$ contains constructed information sets, and $\mathcal{B}_u$ records assigned claim instances.

\begin{algorithm}[t]
\caption{Evaluation Pipeline for Multi-Agent Information Analysis}
\label{alg:information_evaluation}
\footnotesize
\begin{algorithmic}[1]

\STATE \textbf{Input:} $\mathcal{A}$, $x_i^{(t)}$, $S_0$, $\mathcal{H}$, $\{\mathcal{P}_u\}$, $\tau$

\STATE \textbf{Output:} $\mathbf{U}^{(t)}$, $\mathbf{CC}^{(t)}$, $\mathbf{ER}^{(t)}$,
$\mathbf{OR}^{(t)}$, $\mathbf{SR}_{t\rightarrow t'}$, and $\mathbf{SL}(d)$

\STATE \textbf{Construct information units:}

\FOR{each stage $t$}
    \FOR{each agent $a_i\in\mathcal{A}$}
        \STATE $C_i^{(t)}\leftarrow\mathcal{D}(x_i^{(t)})$

        \STATE Compute $R_{\mathrm{src}}(s,c)$ for
        $s\in S_0$ and $c\in C_i^{(t)}$

        \STATE Partition $C_i^{(t)}$ into $S_i^{(t)},E_i^{(t)}$ using $R_{\mathrm{src}}^{*},\tau$

        \STATE Compute $ER_i^{(t)}$

        \STATE Compute $OR_i^{(t)}$ using $P_{\mathrm{OR}}$
    \ENDFOR

    \STATE Aggregate $\{ER_i^{(t)}\}_{i=1}^{N}$
    to obtain $\mathbf{ER}^{(t)}$

    \STATE Aggregate $\{OR_i^{(t)}\}_{i=1}^{N}$
    to obtain $\mathbf{OR}^{(t)}$

    \STATE \textbf{Compute U and CC:}

    \FOR{each agent pair $(a_i,a_j)$, $i<j$}
        \STATE Compute $R(c,d)$ for
        $c\in C_i^{(t)}$ and $d\in C_j^{(t)}$

        \STATE Apply $\tau$ to obtain $\widetilde{R}(c,d)$

        \STATE Compute $\sigma(c)$, $\sigma(d)$, and $A(c,d)$

        \STATE Compute $m(c,d)=\widetilde{R}(c,d)A(c,d)$

        \STATE Compute $U_{a_i\rightarrow a_j}^{(t)}$
        and $U_{a_j\rightarrow a_i}^{(t)}$

        \STATE Compute pairwise Unity $U_{a_i a_j}^{(t)}$
    \ENDFOR

    \STATE Aggregate $\{U_{a_i a_j}^{(t)}\}_{i<j}$
    to obtain $\mathbf{U}^{(t)}$

    \STATE Pool claims into $\mathcal{S}^{(t)}$ and $\mathcal{E}^{(t)}$

    \FOR{each distinct claim pair in $\mathcal{S}^{(t)}$ or $\mathcal{E}^{(t)}$}
        \STATE Compute or reuse $R(c,d)$
    \ENDFOR

    \STATE Cluster $\mathcal{S}^{(t)},\mathcal{E}^{(t)}$ into $\mathcal{K}_{\mathrm{src}}^{(t)},\mathcal{K}_{\mathrm{ext}}^{(t)}$ using $\tau$

    \STATE $\mathbf{CC}^{(t)}\leftarrow|\mathcal{K}_{\mathrm{src}}^{(t)}|+|\mathcal{K}_{\mathrm{ext}}^{(t)}|$
\ENDFOR

\STATE \textbf{Compute cross-stage retention:}

\FOR{each evaluated transition $t\rightarrow t'$}
    \FOR{each agent $a_i\in\mathcal{A}$}
        \FOR{each $e\in E_i^{(t)}$}
            \STATE Compare $e$ with target extensions along the specified propagation path

            \STATE Compute $q_{t\rightarrow t'}(e)$ using $P_{\mathrm{SR}}$
        \ENDFOR
    \ENDFOR

    \STATE Aggregate $\{q_{t\rightarrow t'}(e)\}$
    to obtain $\mathbf{SR}_{t\rightarrow t'}$
\ENDFOR

\STATE \textbf{Construct extension information sets:}

\FOR{each interaction unit $u\in\mathcal{H}$}
    \STATE Build a graph from extension claims ordered by $\mathcal{P}_u$

    \STATE Add same-position cross-agent edges when $R(c,d)\geq\tau$

    \STATE Add adjacent-position edges along the propagation path when $P_{\mathrm{SR}}(c,d)\geq\tau$

    \STATE Initialize $\mathcal{Q}_u,\mathcal{B}_u\leftarrow\varnothing$

    \FOR{each claim $c$ in position order}
        \IF{$c\notin\mathcal{B}_u$}
            \STATE $q\leftarrow$ all reachable, unassigned claims from $c$

            \STATE $\mathcal{Q}_u\leftarrow\mathcal{Q}_u\cup\{q\}$; $\mathcal{B}_u\leftarrow\mathcal{B}_u\cup q$

            \STATE Compute $S(q)=l_q-r_q+1$
        \ENDIF
    \ENDFOR

    \STATE Compute $W_u(d)=\sum_{q\in\mathcal{Q}_u}\mathbb{I}[S(q)\geq d]$
\ENDFOR

\STATE Average $W_u(d)$ over $u\in\mathcal{H}$ to obtain $\overline{W}(d)$

\STATE $\mathbf{SL}(d)\leftarrow\overline{W}(d)/\overline{W}(1)$

\RETURN $\mathbf{U}^{(t)}$, $\mathbf{CC}^{(t)}$, $\mathbf{ER}^{(t)}$,
$\mathbf{OR}^{(t)}$, $\mathbf{SR}_{t\rightarrow t'}$, and $\mathbf{SL}(d)$

\end{algorithmic}
\end{algorithm}
\section{Additional Experimental Results}
\label{app:additional_results}
\begin{figure*}[t]
\centering
\includegraphics[width=\textwidth]{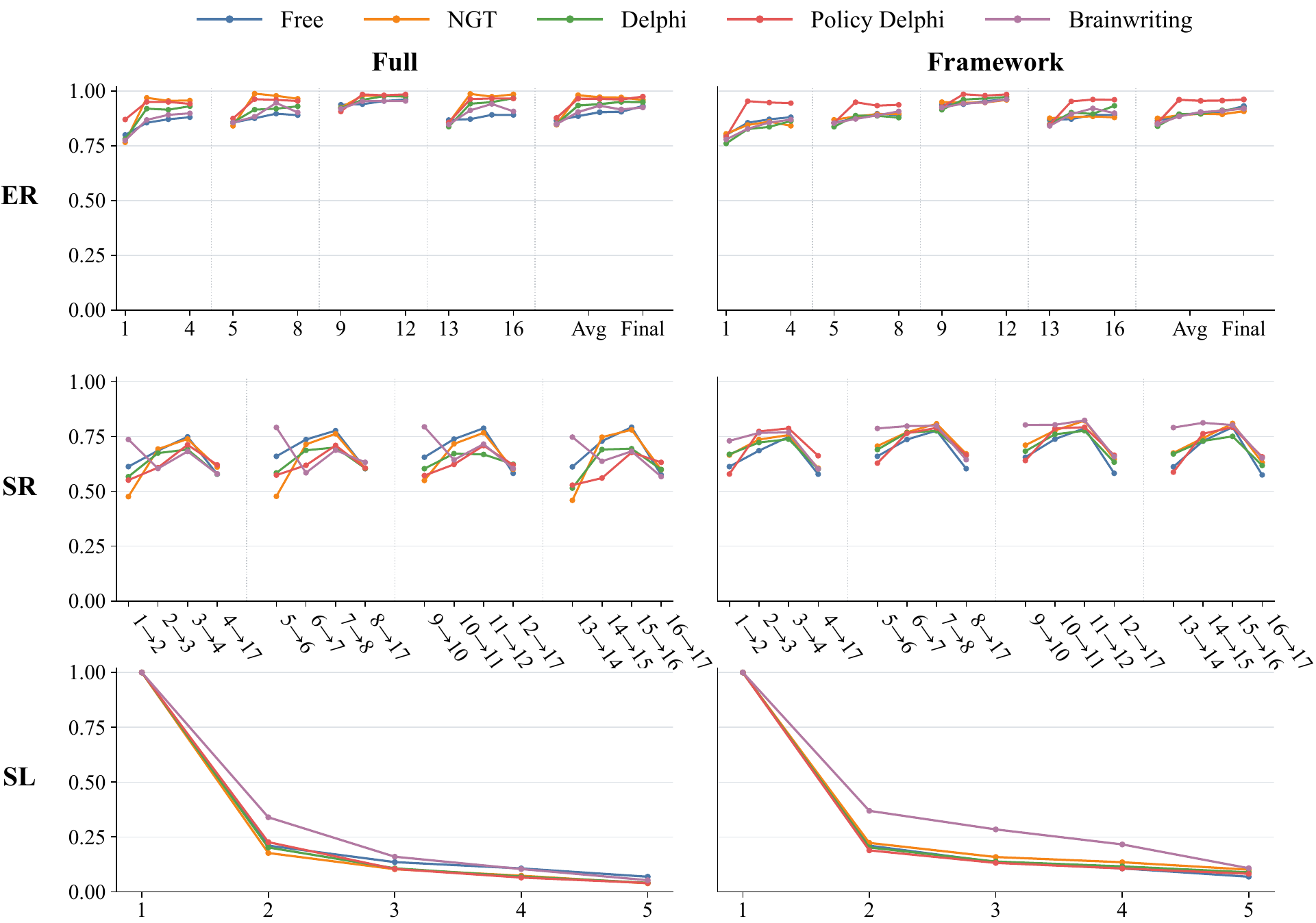}
\caption{Extension Rate, Survival Rate, and extension information-set survival proportions across 40 news cases. Left: Full. Right: Framework. SR includes all four Cycle-endpoint-to-Stage-17 transitions. $SL(d)$ is the proportion of information sets surviving at least $d$ positions under the aggregation in Appendix~\ref{app:aggregation}.}
\label{fig:additional_metrics}
\end{figure*}

Figure~\ref{fig:additional_metrics} complements the main overview in Figure~\ref{fig:metric_overview}. It shows ER at all 17 Stages, SR for all 16 evaluated transitions, and $SL(d)$ for $d=1,\ldots,5$. Colors are shared across the Full and Framework panels. Cycle boundaries are not connected by lines.

The aggregate values for all six metrics are reported in Table~\ref{tab:overall_results}; their calculation is given in Appendix~\ref{app:aggregation}.

\subsection{Numerical Values of the Metric Trajectories}

\label{app:metric_trajectory_values}

Tables~\ref{tab:appendix_u_values}--\ref{tab:appendix_sl_values} report the means underlying the six metric panels. U, CC, ER, and OR include Stages 1--16, the four within-Cycle position averages, and the Stage-17 final summary. Avg-$j$ averages Stages $j$, $j+4$, $j+8$, and $j+12$. SR includes the 16 evaluated transitions, including each Cycle endpoint to Stage 17. SL entries report cumulative lineage survival proportions, not mean SL. Values are rounded only for presentation.

\begin{table*}[t]

\centering

\small

\setlength{\tabcolsep}{4pt}

\textbf{Stages 1--8}\par\smallskip

\begin{tabular}{llcccccccc}
\toprule
\textbf{Strategy} & \textbf{Setting} & \textbf{1} & \textbf{2} & \textbf{3} & \textbf{4} & \textbf{5} & \textbf{6} & \textbf{7} & \textbf{8} \\
\midrule
Free & Control & 0.269 & 0.308 & 0.385 & 0.456 & 0.256 & 0.345 & 0.418 & 0.488 \\
NGT & Full & 0.329 & 0.230 & 0.300 & 0.412 & 0.263 & 0.238 & 0.337 & 0.456 \\
NGT & Framework & 0.417 & 0.458 & 0.535 & 0.606 & 0.387 & 0.504 & 0.631 & 0.670 \\
Delphi & Full & 0.352 & 0.273 & 0.309 & 0.311 & 0.291 & 0.312 & 0.330 & 0.303 \\
Delphi & Framework & 0.443 & 0.485 & 0.490 & 0.574 & 0.419 & 0.588 & 0.599 & 0.651 \\
Policy Delphi & Full & 0.271 & 0.118 & 0.269 & 0.387 & 0.258 & 0.095 & 0.259 & 0.418 \\
Policy Delphi & Framework & 0.416 & 0.450 & 0.560 & 0.592 & 0.396 & 0.465 & 0.575 & 0.618 \\
Brainwriting & Full & 0.316 & 0.204 & 0.152 & 0.199 & 0.265 & 0.187 & 0.106 & 0.164 \\
Brainwriting & Framework & 0.413 & 0.327 & 0.313 & 0.335 & 0.403 & 0.352 & 0.314 & 0.316 \\
\bottomrule
\end{tabular}

\par\medskip

\textbf{Stages 9--16}\par\smallskip

\begin{tabular}{llcccccccc}
\toprule
\textbf{Strategy} & \textbf{Setting} & \textbf{9} & \textbf{10} & \textbf{11} & \textbf{12} & \textbf{13} & \textbf{14} & \textbf{15} & \textbf{16} \\
\midrule
Free & Control & 0.162 & 0.351 & 0.461 & 0.524 & 0.202 & 0.362 & 0.512 & 0.534 \\
NGT & Full & 0.171 & 0.243 & 0.389 & 0.476 & 0.183 & 0.368 & 0.421 & 0.452 \\
NGT & Framework & 0.302 & 0.497 & 0.628 & 0.702 & 0.282 & 0.469 & 0.624 & 0.659 \\
Delphi & Full & 0.205 & 0.222 & 0.222 & 0.240 & 0.191 & 0.301 & 0.325 & 0.285 \\
Delphi & Framework & 0.314 & 0.520 & 0.515 & 0.605 & 0.336 & 0.440 & 0.449 & 0.650 \\
Policy Delphi & Full & 0.216 & 0.101 & 0.266 & 0.384 & 0.236 & 0.073 & 0.222 & 0.356 \\
Policy Delphi & Framework & 0.310 & 0.456 & 0.553 & 0.576 & 0.307 & 0.442 & 0.567 & 0.614 \\
Brainwriting & Full & 0.200 & 0.161 & 0.095 & 0.151 & 0.219 & 0.149 & 0.080 & 0.130 \\
Brainwriting & Framework & 0.314 & 0.272 & 0.262 & 0.269 & 0.320 & 0.255 & 0.216 & 0.223 \\
\bottomrule
\end{tabular}

\par\medskip

\textbf{Position averages and final summary}\par\smallskip

\begin{tabular}{llccccc}
\toprule
\textbf{Strategy} & \textbf{Setting} & \textbf{Avg-1} & \textbf{Avg-2} & \textbf{Avg-3} & \textbf{Avg-4} & \textbf{Final} \\
\midrule
Free & Control & 0.222 & 0.341 & 0.444 & 0.501 & 0.266 \\
NGT & Full & 0.237 & 0.270 & 0.362 & 0.449 & 0.218 \\
NGT & Framework & 0.347 & 0.482 & 0.605 & 0.659 & 0.442 \\
Delphi & Full & 0.260 & 0.277 & 0.296 & 0.285 & 0.207 \\
Delphi & Framework & 0.378 & 0.508 & 0.513 & 0.620 & 0.366 \\
Policy Delphi & Full & 0.245 & 0.097 & 0.254 & 0.386 & 0.219 \\
Policy Delphi & Framework & 0.357 & 0.453 & 0.564 & 0.600 & 0.369 \\
Brainwriting & Full & 0.250 & 0.175 & 0.108 & 0.161 & 0.153 \\
Brainwriting & Framework & 0.362 & 0.301 & 0.276 & 0.286 & 0.227 \\
\bottomrule
\end{tabular}

\caption{Informational Unity (U) across the nine experimental conditions.}

\label{tab:appendix_u_values}

\end{table*}

\begin{table*}[t]

\centering

\small

\setlength{\tabcolsep}{4pt}

\textbf{Stages 1--8}\par\smallskip

\begin{tabular}{llcccccccc}
\toprule
\textbf{Strategy} & \textbf{Setting} & \textbf{1} & \textbf{2} & \textbf{3} & \textbf{4} & \textbf{5} & \textbf{6} & \textbf{7} & \textbf{8} \\
\midrule
Free & Control & 13.65 & 14.38 & 12.32 & 11.05 & 15.43 & 14.55 & 12.40 & 11.32 \\
NGT & Full & 12.00 & 13.75 & 12.95 & 11.10 & 15.05 & 14.93 & 12.78 & 10.40 \\
NGT & Framework & 10.97 & 11.05 & 10.28 & 8.88 & 12.43 & 10.95 & 8.62 & 7.62 \\
Delphi & Full & 11.55 & 14.70 & 13.43 & 13.78 & 14.65 & 14.88 & 14.15 & 14.38 \\
Delphi & Framework & 10.65 & 10.93 & 10.35 & 9.03 & 12.72 & 9.78 & 9.32 & 8.18 \\
Policy Delphi & Full & 18.38 & 17.48 & 14.95 & 11.93 & 17.95 & 19.10 & 15.78 & 12.20 \\
Policy Delphi & Framework & 11.25 & 10.38 & 8.55 & 7.92 & 13.45 & 11.25 & 8.97 & 8.05 \\
Brainwriting & Full & 12.78 & 16.88 & 19.55 & 16.32 & 15.53 & 18.20 & 20.80 & 17.90 \\
Brainwriting & Framework & 11.00 & 14.35 & 13.88 & 12.97 & 12.60 & 13.97 & 14.60 & 13.68 \\
\bottomrule
\end{tabular}

\par\medskip

\textbf{Stages 9--16}\par\smallskip

\begin{tabular}{llcccccccc}
\toprule
\textbf{Strategy} & \textbf{Setting} & \textbf{9} & \textbf{10} & \textbf{11} & \textbf{12} & \textbf{13} & \textbf{14} & \textbf{15} & \textbf{16} \\
\midrule
Free & Control & 18.43 & 15.10 & 12.15 & 11.22 & 14.75 & 12.40 & 10.55 & 9.75 \\
NGT & Full & 17.90 & 15.85 & 12.62 & 10.93 & 13.65 & 11.47 & 10.45 & 9.22 \\
NGT & Framework & 14.57 & 11.82 & 9.53 & 8.10 & 11.47 & 10.65 & 7.78 & 7.28 \\
Delphi & Full & 17.27 & 17.95 & 16.77 & 16.75 & 12.93 & 13.68 & 12.80 & 13.50 \\
Delphi & Framework & 14.68 & 11.57 & 11.45 & 9.50 & 11.07 & 11.15 & 10.85 & 7.55 \\
Policy Delphi & Full & 19.05 & 19.85 & 16.07 & 12.45 & 18.93 & 18.82 & 16.52 & 12.15 \\
Policy Delphi & Framework & 15.70 & 11.60 & 10.07 & 8.97 & 12.68 & 10.03 & 8.62 & 7.38 \\
Brainwriting & Full & 18.00 & 19.38 & 21.15 & 18.88 & 13.22 & 18.50 & 20.70 & 17.30 \\
Brainwriting & Framework & 14.47 & 15.25 & 15.32 & 15.05 & 11.70 & 14.35 & 15.38 & 14.43 \\
\bottomrule
\end{tabular}

\par\medskip

\textbf{Position averages and final summary}\par\smallskip

\begin{tabular}{llccccc}
\toprule
\textbf{Strategy} & \textbf{Setting} & \textbf{Avg-1} & \textbf{Avg-2} & \textbf{Avg-3} & \textbf{Avg-4} & \textbf{Final} \\
\midrule
Free & Control & 15.56 & 14.11 & 11.86 & 10.84 & 19.45 \\
NGT & Full & 14.65 & 14.00 & 12.20 & 10.41 & 19.95 \\
NGT & Framework & 12.36 & 11.12 & 9.05 & 7.97 & 14.90 \\
Delphi & Full & 14.10 & 15.30 & 14.29 & 14.60 & 20.43 \\
Delphi & Framework & 12.28 & 10.86 & 10.49 & 8.56 & 16.18 \\
Policy Delphi & Full & 18.57 & 18.81 & 15.83 & 12.18 & 19.48 \\
Policy Delphi & Framework & 13.27 & 10.81 & 9.06 & 8.08 & 15.60 \\
Brainwriting & Full & 14.88 & 18.24 & 20.55 & 17.60 & 23.98 \\
Brainwriting & Framework & 12.44 & 14.48 & 14.79 & 14.03 & 20.25 \\
\bottomrule
\end{tabular}

\caption{Unique Claim Count (CC) across the nine experimental conditions.}

\label{tab:appendix_cc_values}

\end{table*}

\begin{table*}[t]

\centering

\small

\setlength{\tabcolsep}{4pt}

\textbf{Stages 1--8}\par\smallskip

\begin{tabular}{llcccccccc}
\toprule
\textbf{Strategy} & \textbf{Setting} & \textbf{1} & \textbf{2} & \textbf{3} & \textbf{4} & \textbf{5} & \textbf{6} & \textbf{7} & \textbf{8} \\
\midrule
Free & Control & 0.799 & 0.855 & 0.871 & 0.880 & 0.855 & 0.875 & 0.896 & 0.889 \\
NGT & Full & 0.766 & 0.969 & 0.955 & 0.956 & 0.841 & 0.988 & 0.978 & 0.964 \\
NGT & Framework & 0.805 & 0.845 & 0.860 & 0.841 & 0.868 & 0.884 & 0.894 & 0.896 \\
Delphi & Full & 0.781 & 0.919 & 0.914 & 0.930 & 0.857 & 0.915 & 0.919 & 0.930 \\
Delphi & Framework & 0.760 & 0.826 & 0.836 & 0.863 & 0.836 & 0.887 & 0.886 & 0.878 \\
Policy Delphi & Full & 0.870 & 0.950 & 0.950 & 0.941 & 0.874 & 0.962 & 0.959 & 0.954 \\
Policy Delphi & Framework & 0.790 & 0.953 & 0.947 & 0.944 & 0.854 & 0.948 & 0.933 & 0.937 \\
Brainwriting & Full & 0.774 & 0.868 & 0.891 & 0.898 & 0.856 & 0.883 & 0.945 & 0.902 \\
Brainwriting & Framework & 0.778 & 0.827 & 0.856 & 0.869 & 0.853 & 0.872 & 0.889 & 0.907 \\
\bottomrule
\end{tabular}

\par\medskip

\textbf{Stages 9--16}\par\smallskip

\begin{tabular}{llcccccccc}
\toprule
\textbf{Strategy} & \textbf{Setting} & \textbf{9} & \textbf{10} & \textbf{11} & \textbf{12} & \textbf{13} & \textbf{14} & \textbf{15} & \textbf{16} \\
\midrule
Free & Control & 0.937 & 0.939 & 0.954 & 0.961 & 0.867 & 0.871 & 0.891 & 0.890 \\
NGT & Full & 0.924 & 0.978 & 0.978 & 0.974 & 0.854 & 0.987 & 0.974 & 0.984 \\
NGT & Framework & 0.949 & 0.947 & 0.947 & 0.959 & 0.876 & 0.881 & 0.884 & 0.879 \\
Delphi & Full & 0.922 & 0.959 & 0.976 & 0.975 & 0.837 & 0.942 & 0.949 & 0.967 \\
Delphi & Framework & 0.914 & 0.960 & 0.966 & 0.972 & 0.848 & 0.901 & 0.895 & 0.932 \\
Policy Delphi & Full & 0.906 & 0.984 & 0.980 & 0.984 & 0.856 & 0.963 & 0.966 & 0.966 \\
Policy Delphi & Framework & 0.926 & 0.985 & 0.979 & 0.984 & 0.846 & 0.952 & 0.961 & 0.960 \\
Brainwriting & Full & 0.918 & 0.953 & 0.954 & 0.954 & 0.847 & 0.911 & 0.941 & 0.907 \\
Brainwriting & Framework & 0.924 & 0.942 & 0.948 & 0.962 & 0.841 & 0.894 & 0.920 & 0.899 \\
\bottomrule
\end{tabular}

\par\medskip

\textbf{Position averages and final summary}\par\smallskip

\begin{tabular}{llccccc}
\toprule
\textbf{Strategy} & \textbf{Setting} & \textbf{Avg-1} & \textbf{Avg-2} & \textbf{Avg-3} & \textbf{Avg-4} & \textbf{Final} \\
\midrule
Free & Control & 0.865 & 0.885 & 0.903 & 0.905 & 0.931 \\
NGT & Full & 0.846 & 0.980 & 0.971 & 0.970 & 0.959 \\
NGT & Framework & 0.875 & 0.889 & 0.896 & 0.894 & 0.908 \\
Delphi & Full & 0.849 & 0.934 & 0.940 & 0.950 & 0.949 \\
Delphi & Framework & 0.840 & 0.894 & 0.896 & 0.911 & 0.920 \\
Policy Delphi & Full & 0.877 & 0.965 & 0.964 & 0.961 & 0.974 \\
Policy Delphi & Framework & 0.854 & 0.960 & 0.955 & 0.956 & 0.961 \\
Brainwriting & Full & 0.849 & 0.904 & 0.933 & 0.915 & 0.925 \\
Brainwriting & Framework & 0.849 & 0.884 & 0.903 & 0.909 & 0.920 \\
\bottomrule
\end{tabular}

\caption{Extension Rate (ER) across the nine experimental conditions.}

\label{tab:appendix_er_values}

\end{table*}

\begin{table*}[t]

\centering

\small

\setlength{\tabcolsep}{4pt}

\textbf{Stages 1--8}\par\smallskip

\begin{tabular}{llcccccccc}
\toprule
\textbf{Strategy} & \textbf{Setting} & \textbf{1} & \textbf{2} & \textbf{3} & \textbf{4} & \textbf{5} & \textbf{6} & \textbf{7} & \textbf{8} \\
\midrule
Free & Control & 0.815 & 0.794 & 0.767 & 0.760 & 0.789 & 0.759 & 0.731 & 0.723 \\
NGT & Full & 0.841 & 0.432 & 0.557 & 0.575 & 0.804 & 0.366 & 0.451 & 0.448 \\
NGT & Framework & 0.823 & 0.817 & 0.818 & 0.809 & 0.783 & 0.770 & 0.779 & 0.777 \\
Delphi & Full & 0.848 & 0.692 & 0.694 & 0.640 & 0.787 & 0.673 & 0.669 & 0.665 \\
Delphi & Framework & 0.851 & 0.819 & 0.817 & 0.788 & 0.812 & 0.771 & 0.763 & 0.762 \\
Policy Delphi & Full & 0.816 & 0.512 & 0.564 & 0.576 & 0.818 & 0.442 & 0.533 & 0.558 \\
Policy Delphi & Framework & 0.845 & 0.658 & 0.686 & 0.691 & 0.821 & 0.668 & 0.699 & 0.716 \\
Brainwriting & Full & 0.848 & 0.801 & 0.725 & 0.769 & 0.796 & 0.772 & 0.680 & 0.722 \\
Brainwriting & Framework & 0.851 & 0.808 & 0.792 & 0.803 & 0.788 & 0.762 & 0.758 & 0.761 \\
\bottomrule
\end{tabular}

\par\medskip

\textbf{Stages 9--16}\par\smallskip

\begin{tabular}{llcccccccc}
\toprule
\textbf{Strategy} & \textbf{Setting} & \textbf{9} & \textbf{10} & \textbf{11} & \textbf{12} & \textbf{13} & \textbf{14} & \textbf{15} & \textbf{16} \\
\midrule
Free & Control & 0.706 & 0.698 & 0.692 & 0.684 & 0.630 & 0.621 & 0.607 & 0.600 \\
NGT & Full & 0.732 & 0.432 & 0.527 & 0.519 & 0.647 & 0.233 & 0.269 & 0.271 \\
NGT & Framework & 0.749 & 0.743 & 0.759 & 0.762 & 0.690 & 0.708 & 0.726 & 0.709 \\
Delphi & Full & 0.723 & 0.594 & 0.559 & 0.511 & 0.657 & 0.471 & 0.471 & 0.383 \\
Delphi & Framework & 0.776 & 0.720 & 0.729 & 0.691 & 0.737 & 0.660 & 0.656 & 0.607 \\
Policy Delphi & Full & 0.759 & 0.383 & 0.435 & 0.470 & 0.738 & 0.335 & 0.389 & 0.382 \\
Policy Delphi & Framework & 0.785 & 0.633 & 0.650 & 0.634 & 0.735 & 0.550 & 0.546 & 0.556 \\
Brainwriting & Full & 0.741 & 0.677 & 0.614 & 0.651 & 0.654 & 0.571 & 0.459 & 0.553 \\
Brainwriting & Framework & 0.769 & 0.735 & 0.722 & 0.735 & 0.714 & 0.658 & 0.613 & 0.648 \\
\bottomrule
\end{tabular}

\par\medskip

\textbf{Position averages and final summary}\par\smallskip

\begin{tabular}{llccccc}
\toprule
\textbf{Strategy} & \textbf{Setting} & \textbf{Avg-1} & \textbf{Avg-2} & \textbf{Avg-3} & \textbf{Avg-4} & \textbf{Final} \\
\midrule
Free & Control & 0.735 & 0.718 & 0.699 & 0.692 & 0.749 \\
NGT & Full & 0.756 & 0.366 & 0.451 & 0.453 & 0.614 \\
NGT & Framework & 0.761 & 0.759 & 0.770 & 0.764 & 0.795 \\
Delphi & Full & 0.754 & 0.608 & 0.598 & 0.550 & 0.682 \\
Delphi & Framework & 0.794 & 0.742 & 0.741 & 0.712 & 0.792 \\
Policy Delphi & Full & 0.782 & 0.418 & 0.480 & 0.496 & 0.581 \\
Policy Delphi & Framework & 0.796 & 0.627 & 0.645 & 0.649 & 0.730 \\
Brainwriting & Full & 0.760 & 0.705 & 0.619 & 0.674 & 0.749 \\
Brainwriting & Framework & 0.780 & 0.741 & 0.721 & 0.737 & 0.766 \\
\bottomrule
\end{tabular}

\caption{Original News Retention (OR) across the nine experimental conditions.}

\label{tab:appendix_or_values}

\end{table*}

\begin{table*}[t]

\centering

\small

\setlength{\tabcolsep}{4pt}

\textbf{Cycles 1--2}\par\smallskip

\begin{tabular}{llcccccccc}
\toprule
\textbf{Strategy} & \textbf{Setting} & \textbf{$1\!\to\!2$} & \textbf{$2\!\to\!3$} & \textbf{$3\!\to\!4$} & \textbf{$4\!\to\!17$} & \textbf{$5\!\to\!6$} & \textbf{$6\!\to\!7$} & \textbf{$7\!\to\!8$} & \textbf{$8\!\to\!17$} \\
\midrule
Free & Control & 0.613 & 0.686 & 0.749 & 0.578 & 0.660 & 0.736 & 0.777 & 0.603 \\
NGT & Full & 0.476 & 0.693 & 0.739 & 0.611 & 0.477 & 0.714 & 0.762 & 0.604 \\
NGT & Framework & 0.665 & 0.736 & 0.757 & 0.606 & 0.707 & 0.769 & 0.808 & 0.671 \\
Delphi & Full & 0.567 & 0.674 & 0.691 & 0.579 & 0.584 & 0.687 & 0.700 & 0.605 \\
Delphi & Framework & 0.669 & 0.723 & 0.738 & 0.601 & 0.690 & 0.768 & 0.776 & 0.664 \\
Policy Delphi & Full & 0.551 & 0.608 & 0.712 & 0.621 & 0.574 & 0.619 & 0.709 & 0.607 \\
Policy Delphi & Framework & 0.579 & 0.773 & 0.787 & 0.662 & 0.629 & 0.765 & 0.789 & 0.666 \\
Brainwriting & Full & 0.736 & 0.604 & 0.682 & 0.580 & 0.791 & 0.585 & 0.688 & 0.632 \\
Brainwriting & Framework & 0.730 & 0.767 & 0.769 & 0.602 & 0.787 & 0.798 & 0.799 & 0.644 \\
\bottomrule
\end{tabular}

\par\medskip

\textbf{Cycles 3--4}\par\smallskip

\begin{tabular}{llcccccccc}
\toprule
\textbf{Strategy} & \textbf{Setting} & \textbf{$9\!\to\!10$} & \textbf{$10\!\to\!11$} & \textbf{$11\!\to\!12$} & \textbf{$12\!\to\!17$} & \textbf{$13\!\to\!14$} & \textbf{$14\!\to\!15$} & \textbf{$15\!\to\!16$} & \textbf{$16\!\to\!17$} \\
\midrule
Free & Control & 0.656 & 0.738 & 0.788 & 0.582 & 0.612 & 0.729 & 0.792 & 0.575 \\
NGT & Full & 0.550 & 0.716 & 0.767 & 0.601 & 0.459 & 0.747 & 0.780 & 0.599 \\
NGT & Framework & 0.710 & 0.778 & 0.822 & 0.648 & 0.675 & 0.742 & 0.809 & 0.631 \\
Delphi & Full & 0.603 & 0.672 & 0.668 & 0.624 & 0.515 & 0.691 & 0.695 & 0.600 \\
Delphi & Framework & 0.683 & 0.760 & 0.776 & 0.633 & 0.670 & 0.730 & 0.751 & 0.618 \\
Policy Delphi & Full & 0.572 & 0.623 & 0.708 & 0.620 & 0.528 & 0.561 & 0.677 & 0.632 \\
Policy Delphi & Framework & 0.641 & 0.787 & 0.793 & 0.665 & 0.588 & 0.763 & 0.792 & 0.657 \\
Brainwriting & Full & 0.794 & 0.644 & 0.716 & 0.604 & 0.747 & 0.637 & 0.682 & 0.567 \\
Brainwriting & Framework & 0.803 & 0.803 & 0.824 & 0.658 & 0.791 & 0.812 & 0.802 & 0.652 \\
\bottomrule
\end{tabular}

\caption{Survival Rate (SR) across the nine experimental conditions.}

\label{tab:appendix_sr_values}

\end{table*}

\subsection{Mean Correlations Among the Six Metrics}

\label{app:metric_correlations}

Table~\ref{tab:appendix_corr_mean} reports the arithmetic means of the nine setting-specific correlation coefficients; observations are not pooled. Pairs among U, CC, ER, and OR use the 17 Stage-level means across 40 cases. For SR pairs excluding SL, the other metric is aligned to each transition destination, including four occurrences of Stage 17. All pairs involving SL use 40 case-level scalar summaries. Each cell reports Pearson on the first line and Spearman on the second. The matrix is symmetric, with self-correlations of 1 on the diagonal.

\twocolumn[{%
\begin{minipage}{\textwidth}
\expandafter\def\csname @captype\endcsname{table}
\begingroup
\centering

\small

\setlength{\tabcolsep}{4pt}

\begin{tabular}{llccccc}
\toprule
\textbf{Strategy} & \textbf{Setting} & \textbf{$\mathrm{SL}\geq 1$} & \textbf{$\mathrm{SL}\geq 2$} & \textbf{$\mathrm{SL}\geq 3$} & \textbf{$\mathrm{SL}\geq 4$} & \textbf{$\mathrm{SL}\geq 5$} \\
\midrule
Free & Control & 1.000 & 0.210 & 0.135 & 0.107 & 0.069 \\
NGT & Full & 1.000 & 0.177 & 0.103 & 0.073 & 0.038 \\
NGT & Framework & 1.000 & 0.223 & 0.158 & 0.135 & 0.101 \\
Delphi & Full & 1.000 & 0.202 & 0.107 & 0.069 & 0.041 \\
Delphi & Framework & 1.000 & 0.202 & 0.138 & 0.116 & 0.089 \\
Policy Delphi & Full & 1.000 & 0.226 & 0.104 & 0.065 & 0.040 \\
Policy Delphi & Framework & 1.000 & 0.189 & 0.132 & 0.106 & 0.082 \\
Brainwriting & Full & 1.000 & 0.340 & 0.160 & 0.104 & 0.053 \\
Brainwriting & Framework & 1.000 & 0.369 & 0.284 & 0.216 & 0.107 \\
\bottomrule
\end{tabular}

\caption{Claim Survival Length (SL) across the nine experimental conditions. Each entry is the proportion of extension lineages surviving at least the indicated number of positions.}

\label{tab:appendix_sl_values}
\endgroup

\par\vspace{14pt}

\begingroup
\centering

\small

\setlength{\tabcolsep}{8pt}

\renewcommand{\arraystretch}{1.05}
\setlength{\cmidrulesep}{1pt}
\setlength{\aboverulesep}{1.5pt}
\setlength{\belowrulesep}{1.5pt}

\begin{tabular}{lcccccc}
\toprule
\textbf{Metric} & \textbf{U} & \textbf{CC} & \textbf{ER} & \textbf{OR} & \textbf{SR} & \textbf{SL} \\
\midrule
\multirow{2}{*}{\textbf{U}} & $1.000$ & $-0.815$ & $-0.048$ & $0.040$ & $0.715$ & $0.222$ \\
\cmidrule(lr){2-2}\cmidrule(lr){3-3}\cmidrule(lr){4-4}\cmidrule(lr){5-5}\cmidrule(lr){6-6}\cmidrule(lr){7-7}
 & $1.000$ & $-0.825$ & $-0.038$ & $0.016$ & $0.650$ & $0.229$ \\
\midrule
\multirow{2}{*}{\textbf{CC}} & $-0.815$ & $1.000$ & $0.173$ & $0.156$ & $-0.710$ & $-0.232$ \\
\cmidrule(lr){2-2}\cmidrule(lr){3-3}\cmidrule(lr){4-4}\cmidrule(lr){5-5}\cmidrule(lr){6-6}\cmidrule(lr){7-7}
 & $-0.825$ & $1.000$ & $0.182$ & $0.105$ & $-0.661$ & $-0.253$ \\
\midrule
\multirow{2}{*}{\textbf{ER}} & $-0.048$ & $0.173$ & $1.000$ & $-0.634$ & $-0.037$ & $0.166$ \\
\cmidrule(lr){2-2}\cmidrule(lr){3-3}\cmidrule(lr){4-4}\cmidrule(lr){5-5}\cmidrule(lr){6-6}\cmidrule(lr){7-7}
 & $-0.038$ & $0.182$ & $1.000$ & $-0.666$ & $0.008$ & $0.202$ \\
\midrule
\multirow{2}{*}{\textbf{OR}} & $0.040$ & $0.156$ & $-0.634$ & $1.000$ & $-0.155$ & $0.138$ \\
\cmidrule(lr){2-2}\cmidrule(lr){3-3}\cmidrule(lr){4-4}\cmidrule(lr){5-5}\cmidrule(lr){6-6}\cmidrule(lr){7-7}
 & $0.016$ & $0.105$ & $-0.666$ & $1.000$ & $-0.256$ & $0.143$ \\
\midrule
\multirow{2}{*}{\textbf{SR}} & $0.715$ & $-0.710$ & $-0.037$ & $-0.155$ & $1.000$ & $0.533$ \\
\cmidrule(lr){2-2}\cmidrule(lr){3-3}\cmidrule(lr){4-4}\cmidrule(lr){5-5}\cmidrule(lr){6-6}\cmidrule(lr){7-7}
 & $0.650$ & $-0.661$ & $0.008$ & $-0.256$ & $1.000$ & $0.533$ \\
\midrule
\multirow{2}{*}{\textbf{SL}} & $0.222$ & $-0.232$ & $0.166$ & $0.138$ & $0.533$ & $1.000$ \\
\cmidrule(lr){2-2}\cmidrule(lr){3-3}\cmidrule(lr){4-4}\cmidrule(lr){5-5}\cmidrule(lr){6-6}\cmidrule(lr){7-7}
 & $0.229$ & $-0.253$ & $0.202$ & $0.143$ & $0.533$ & $1.000$ \\
\bottomrule
\end{tabular}

\caption{Mean metric correlations across the nine experimental conditions. Each cell reports Pearson (top) and Spearman (bottom).}

\label{tab:appendix_corr_mean}
\endgroup
\par\vspace{8pt}
\end{minipage}
}]

\section{Human Validation Details}
For DBS-eval and Polis, we use \texttt{gemini-2.5-flash-lite}
for semantic judgments in computing Informational Unity.
For DELPHISTAR, Unity is computed directly from participants'
ratings of the same predefined item.
\label{app:human_validation}
\subsection{DBS-eval}
\label{app:dbs}

DBS-eval provides human-written summaries and annotations of
Summary Content Units (SCUs) \citep{tauchmann2020dbseval}.
Each topic comprises a collection of source documents and
human-written summaries of that collection.
We use all ten topics listed in Table 1 of the original paper.
Each topic contains three or four human-written summaries,
as described in Section 3.2 of that paper.
Each topic forms one validation case, with the complete summaries
treated as individual participant outputs.
Gold SCUs and reference scores are excluded from evaluator inputs.

The reference is the author-reported mean SCU weight (Pyr).
An SCU's weight counts the summaries containing that unit,
so higher mean weights indicate more shared semantic content.
We use the reported values directly to compute correlations with Unity.

For the Simple baseline, an LLM assigns one holistic U score
to all complete participant summaries for each topic.
Both formal U and Simple are correlated with the same reference
scores, as reported in Table~\ref{tab:unity_simple}.
The three-evaluator comparison uses the same ten topics and
fixes the decomposed claims while varying the semantic judge
(Table~\ref{tab:evaluator_consistency}).

\paragraph{Direct LLM Scoring Prompt.}
The prompt is abbreviated below.
\begin{Verbatim}[fontsize=\scriptsize,breaklines=true]
You are evaluating complete human-written summaries.
Unity is the degree of shared specific information and aligned stances.
Give one holistic score in [0,1] from {complete_summaries},
without claim decomposition.
Return JSON: {"unity": 0.0, "confidence": 0.0, "reason": "..."}
\end{Verbatim}
\subsection{DELPHISTAR}
\label{app:delphistar}

DELPHISTAR contains participant ratings of reporting items
in a Delphi study \citep{niederberger2024delphi}.
We evaluate 33 items from the Round-3 data.
Each item forms one validation case, with participants'
valid importance ratings on a 1--7 scale treated as
individual responses.

Because participants rate the same predefined item,
claim correspondence is established by item identity.
We map ratings linearly to the range [0,1] and calculate
U as the mean pairwise stance agreement.

The reference score is the proportion of valid ratings equal to 6 or 7 for each item, excluding missing ratings from the denominator. We correlate U directly with this continuous proportion. 
\subsection{Polis}
\label{app:polis}

Polis records participant votes on identified statements
\citep{small2021polis}.
We evaluate 16 completed derived cases, each containing a cohort of 12 participants and a selected
statement set.
Agree votes retain the corresponding statement,
disagree votes produce an explicit disagreement expression,
and pass or missing votes contribute no claim.
These vote-derived expressions form the participant inputs.

High cases require at least ten of the twelve participants
to agree with each selected statement, together with an
overall agreement rate of at least 80\% in the source data.
Low cases require at least 80\% non-pass vote coverage,
with the minority comprising at least one third of
the agree/disagree votes for each selected statement.
Each participant has at least eight non-pass observations.
High and Low are coded as 1 and 0, respectively.

\subsection{Correlation Analysis}
\label{app:validation_statistics}

Within each dataset, Pearson and Spearman correlations
measure linear and rank associations between U and the
corresponding reference scores.
The sample sizes are ten topics for DBS-eval,
33 items for DELPHISTAR, and 16 derived cases for Polis.
We retain the continuous reference scores for DBS-eval
and DELPHISTAR; Polis uses the predefined binary
High/Low proxies. U remains continuous in all analyses.
Spearman correlations are computed using average ranks
for tied values.
Table~\ref{tab:unity_validation} reports the correlations,
and Tables~\ref{tab:unity_simple}
and~\ref{tab:evaluator_consistency} report the
DBS baseline and evaluator comparisons.
\section{News Sources}
\label{app:news_sources}
The 40 experimental news inputs are constructed from 20 Reuters reports. Each report is used to create one Event-only input and one Event+Seed input. The experimental inputs concisely paraphrase the factual content of the corresponding reports. The original source articles are listed below.

\small

\begin{enumerate}

\item \textbf{Indonesians climb over logs in walk to aid centre as flood deaths exceed 900.}
Reuters, December 6, 2025.\\
\url{https://www.reuters.com/business/environment/indonesians-climb-over-logs-walk-aid-centre-flood-deaths-rise-over-900-2025-12-06/}

\item \textbf{Myanmar faces humanitarian crisis after quake, IFRC says.}
Reuters, March 30, 2025.\\
\url{https://www.reuters.com/business/environment/myanmar-faces-humanitarian-crisis-after-quake-ifrc-says-2025-03-30/}

\item \textbf{Flood-hit central Europe will get billions in EU aid, von der Leyen promises.}
Reuters, September 20, 2024.\\
\url{https://www.reuters.com/world/europe/polands-third-largest-city-braces-peaking-floods-2024-09-19/}

\item \textbf{Google commits \$1 billion for AI training at US universities.}
Reuters, August 6, 2025.\\
\url{https://www.reuters.com/world/us/google-commits-1-billion-ai-training-us-universities-2025-08-06/}

\item \textbf{Chicago Law bans phones and laptops from first-year classes to combat AI.}
Reuters, July 9, 2026.\\
\url{https://www.reuters.com/legal/legalindustry/chicago-law-bans-phones-laptops-first-year-classes-combat-ai-2026-07-09/}

\item \textbf{US FDA approves first mRNA flu shot from Moderna.}
Reuters, August 6, 2026.\\
\url{https://www.reuters.com/business/healthcare-pharmaceuticals/us-fda-approves-first-mrna-flu-shot-moderna-2026-08-06/}

\item \textbf{US agency reviews school internet subsidies amid concerns about kids' screen time.}
Reuters, June 3, 2026.\\
\url{https://www.reuters.com/technology/us-agency-considers-reforming-ending-3-billion-school-internet-subsidy-program-2026-06-03/}

\item \textbf{Thomson Reuters to cut 'small number' of engineering jobs.}
Reuters, July 13, 2026.\\
\url{https://www.reuters.com/legal/litigation/thomson-reuters-cut-small-number-engineering-jobs-2026-07-13/}

\item \textbf{Blue Origin's New Glenn rocket deploys Mars satellites, sticks booster landing.}
Reuters, November 13, 2025.\\
\url{https://www.reuters.com/science/blue-origin-launches-new-glenn-rocket-companys-first-nasa-scale-science-mission-2025-11-13/}

\item \textbf{China launches mission to retrieve asteroid samples.}
Reuters, May 29, 2025 (current article timestamp; URL dated May 28).\\
\url{https://www.reuters.com/science/china-launches-mission-retrieve-asteroid-samples-2025-05-28/}

\item \textbf{Norway will speed up onshore wind power developments, government says.}
Reuters, August 7, 2026.\\
\url{https://www.reuters.com/business/energy/norway-will-speed-up-onshore-wind-power-developments-government-says-2026-08-07/}

\item \textbf{US to launch AI and cybersecurity coordination group, White House says.}
Reuters, July 14, 2026.\\
\url{https://www.reuters.com/technology/us-launch-ai-cybersecurity-coordination-group-white-house-says-2026-07-14/}

\item \textbf{WADA condemns 'dangerous and irresponsible' Enhanced Games.}
Reuters, May 22, 2025.\\
\url{https://www.reuters.com/sports/wada-condemns-dangerous-irresponsible-enhanced-games-2025-05-22/}

\item \textbf{Indian tribes visit UK museum to bring home ancestors' remains.}
Reuters, June 16, 2025.\\
\url{https://www.reuters.com/world/india/indian-tribes-visit-uk-museum-bring-home-ancestors-remains-2025-06-16/}

\item \textbf{BHP faces weekend strike at Port Hedland iron ore hub as wage talks drag.}
Reuters, August 4, 2026.\\
\url{https://www.reuters.com/business/world-at-work/bhp-faces-weekend-strike-port-hedland-iron-ore-hub-wage-talks-drag-2026-08-04/}

\item \textbf{Japan food self-sufficiency ratio falls to 37\%, below 2030 goal of 45\%.}
Reuters, August 7, 2026.\\
\url{https://www.reuters.com/world/asia-pacific/japan-food-self-sufficiency-ratio-falls-37-below-2030-goal-45-2026-08-07/}

\item \textbf{Uber rebrands 'Green' as 'Electric', offers US drivers \$4,000 to switch to EVs.}
Reuters, October 22, 2025.\\
\url{https://www.reuters.com/business/autos-transportation/uber-rebrands-green-electric-offers-us-drivers-4000-switch-evs-2025-10-22/}

\item \textbf{US colleges see 17\% drop in newly enrolled international students, report finds.}
Reuters, November 17, 2025.\\
\url{https://www.reuters.com/world/us/us-colleges-see-17-drop-newly-enrolled-international-students-report-finds-2025-11-17/}

\item \textbf{Countries adopt multibillion-dollar fund to protect nature.}
Reuters, August 24, 2023 (initial report; the current article was updated August 25).\\
\url{https://www.reuters.com/business/environment/countries-launch-fund-protect-nature-un-calls-more-money-2023-08-24/}

\item \textbf{Mexican jalapenos tied to Salmonella outbreak in 27 US states.}
Reuters, August 5, 2026.\\
\url{https://www.reuters.com/business/healthcare-pharmaceuticals/cdc-says-345-sickened-salmonella-outbreak-tied-jalapeno-peppers-2026-08-05/}

\end{enumerate}
\end{document}